\documentclass[journal]{IEEEtran}
\usepackage{amsmath,amsthm,amssymb,amsfonts}
\usepackage{algorithm,algorithmic}
\usepackage{graphicx}
\usepackage{textcomp}
\usepackage{hyperref}
\usepackage{xcolor}
\usepackage{cite}
\usepackage{bm}
\usepackage{pifont}

\def\BibTeX{{\rm B\kern-.05em{\sc i\kern-.025em b}\kern-.08em
    T\kern-.1667em\lower.7ex\hbox{E}\kern-.125emX}}

\makeatletter
\newcommand{\HEADER}[1]{\ALC@it\underline{\textsc{#1}}\begin{ALC@g}}
\newcommand{\ENDHEADER}{\end{ALC@g}}
\makeatother

\newtheorem{theorem}{Theorem}
\newtheorem{prop}{Proposition}
\newtheorem{lemma}{Lemma}

\def\reals{\mathbb{R}} 
\def\E{\mathbb{E}} 
\def\P{\mathbb{P}} 
\def\Gsn{\mathcal{N}}
\newcommand{\trans}[1]{{#1}^{\top}}
\newcommand{\invtrans}[1]{{#1}^{-\top}}
\newcommand{\iid}{\stackrel{\mathrm{i.i.d.}}{\sim}}
\DeclareMathOperator*{\argmax}{arg\,max}
\DeclareMathOperator*{\argmin}{arg\,min}

\newcommand{\cmark}{\ding{51}}
\newcommand{\xmark}{\ding{55}}

\begin{document}
\title{Asymmetric Graph Error Control with Low Complexity in Causal Bandits}
\author{Chen Peng, Di Zhang and Urbashi Mitra
\thanks{This work is funded in part by one or more of the following grants: ARO W911NF1910269, ARO W911NF2410094, DOE DE-SC0021417, Swedish Research Council 2018-04359, NSF A22-2666-S003, NSF A25-0139-S009, NSF CCF-2008927, NSF CCF-2200221, NSF CCF-2311653, NSF KR-705319, NSF RINGS-2148313, NSF PIPP phase II: Center for Pandemic Insights (CPI) 2412522, ONR 503400-78050, ONR N00014-22-1-2363. }
\thanks{Parts of this paper were presented in part at the 2024 IEEE International Conference on Acoustics, Speech and Signal Processing (ICASSP) \cite{peng2024graph}.}
}

\maketitle

\begin{abstract}
In this paper, the causal bandit problem is investigated, with the objective of maximizing the long-term reward by selecting an optimal sequence of interventions on nodes in an unknown causal graph. It is assumed that both the causal topology and the distribution of interventions are unknown. First, based on the difference between the two types of graph identification errors (false positives and negatives), a causal graph learning method is proposed. Numerical results suggest that this method has a much lower sample complexity relative to the prior art by learning \emph{sub-graphs}. However, we note that a sample complexity analysis for the new algorithm has not been undertaken, as of yet. Under the assumption of minimum-mean squared error weight estimation, a new uncertainty bound tailored to the causal bandit problem is derived. This uncertainty bound drives an upper confidence bound-based intervention selection to optimize the reward. Further, we consider a particular instance of non-stationary bandits wherein both the causal topology and interventional distributions can change. Our solution is the design of a sub-graph change detection mechanism that requires a modest number of samples. Numerical results compare the new methodology to existing schemes and show a substantial performance improvement in stationary and non-stationary settings. Averaged over $100$ randomly generated causal bandits, the proposed scheme takes significantly fewer samples to learn the causal structure and achieves a reward gain of $85\%$ compared to existing approaches.
\end{abstract}


\section{Introduction}
Investigation of cause-effect relationships is central to answering key questions in medicine, epidemiology, economics as well as the social and behavioral sciences (see \textit{e.g.} \cite{pearl2009causality,wu2013algorithms,ogburn2024causal}). Cause-effect relationships can be represented by Bayesian networks, in the form of directed acyclic graphs (DAGs). Identifying the causal structural model is essential to decision making and answering counterfactual questions. Herein, we will study methods for determining the underlying causal structural model in the context of causal bandits and thus will also optimize intervention design. We first discuss causal graph identification and then discuss causal bandits.

\subsection{Related Work}
Current causal graph identification methods mainly fall into two categories: constraint-based and score-based. Constraint-based approaches, such as\cite{pearl2009causality,spirtes2001causation}, search for conditional independence over all possible graphs and thus computation complexity scales exponentially. Classical score-based methods evaluate different graph structures in their ability to fit the data (see \textit{e.g.} \cite{chickering2002optimal,meinshausen2006high,koivisto2006advances}). To lower the complexity, recent score-based methods (see \textit{e.g.} \cite{zheng2018dags,ng2020role,bello2022dagma,cui2024topology}) formulate the structure learning problem as a continuous optimization task. In \cite{ng2020role}, an algorithm called GOLEM is proposed, where the likelihood score is combined with a soft DAG constraint to evaluate possible causal structures. The DAGMA scheme proposed in \cite{bello2022dagma} characterizes acyclicity by a log-determinant function, with better-behaved gradients.

Decision-making or action-taking is natural within the context of causal inference. Despite this natural relationship, incorporation of causal methods within the context of multi-armed bandits (MAB) \cite{bareinboim2015bandits,lattimore2016causal}, has only occurred recently.  MABs provide a useful model for sequential decision-making problems, and as such have been employed in the design of clinical trials \cite{liu2020reinforcement}, recommendation systems \cite{xu2016personalized}, financial portfolio design \cite{cai2019gaussian}, \textit{etc}. In a MAB, an agent selects an action (arm) in each round and observes corresponding outcomes -- the goal is to  maximize the long-term cumulative reward. In the standard setting, the stochastic rewards generated by different arms are assumed to be statistically independent. 

To model realistic scenarios with dependence, structured bandits have been considered. Within this arena, we focus on \textbf{causal} bandits \cite{lattimore2016causal}, where the causal structure is exploited to improve decision-making. DAGs can encode the causal relationship among factors that contribute to the reward \cite{lattimore2016causal}. Herein, we interpret the arms as different interventions on the nodes of a DAG and the reward as the stochastic outcome of a certain node. In the current work, we will provide new methods for both causal discovery (graph identification) as well as the design of optimal interventions for causal MABs.

\begin{table*}[t]
\centering
\caption{Key Assumptions: Related Work}
\label{tab:related}
\begin{tabular}{|cccccc|}
\hline
Reference & Topology Knowledge & Intervention & Structural Model & Non-stationary Structure & Exogenous/Noise \\
\hline
\cite{lattimore2016causal,nair2021budgeted,maiti2022causal} & \cmark & Hard & Discrete & \xmark & Bernoulli \\
\hline
\cite{varici2023causal} & \cmark & Soft & Linear & \xmark & Sub-Gaussian \\
\hline
\cite{sussex2022model} & \cmark & Soft & General & \xmark & Known distribution \\
\hline
\cite{yan2024causal} & \cmark & Soft & General & \xmark & Sub-Gaussian \\
\hline
\cite{de2022causal} & Known Separating Sets & Hard & Discrete \& Linear & \xmark & Bernoulli \& Gaussian \\
\hline
\cite{lu2021causal} & \xmark & Hard & Discrete & \xmark & Sub-Gaussian \\
\hline
\cite{malek2023additive} & \xmark & Hard & Linear & \xmark & sub-Gaussian \\
\hline
\cite{yan2024robust} & \cmark & Soft & Linear & \cmark & sub-Gaussian \\
\hline
This work & \xmark & Soft & Linear & \cmark & Gaussian \\
\hline
\end{tabular}
\end{table*}

Existing literature on causal bandits can be categorized based on their assumptions about the topology of the underlying DAG and the distribution of the interventions. While many works assume prior topology or DAG knowledge \cite{lattimore2016causal,nair2021budgeted,maiti2022causal,sussex2022model,varici2023causal,yan2024causal}, this is often not known in practice. In this work, we assume that the underlying causal structure is unknown.

The setting without the knowledge of both topology and interventional distribution has been investigated recently \cite{lu2021causal,de2022causal,malek2023additive,bilodeau2022adaptively}. An algorithm based on causal tree recovery is proposed in \cite{lu2021causal}, with regret scaling logarithmically with the number of nodes. In \cite{de2022causal}, an auxiliary separating set algorithm is utilized to learn the causal structure and provide improved regret over non-causal algorithms. The existing works are based on the hard (perfect) intervention model, where the causal relations between a node and its parents are completely cut off because a specific value is assigned to the node. Herein, we consider soft interventions, such that the node under intervention can still be causally related to other nodes. 

A challenge in MAB problems is the balance between exploration and exploitation. A common approach is via derivations of upper confidence bounds (UCB) \cite{tekin2018multi}. The classic UCB scheme uses the number of visits as a general measure of uncertainty \cite{sutton2018reinforcement}, while in causal bandits, uncertainty can be better quantified by bounding the variance of problem-specific estimators \cite{varici2023causal,lu2021causal,de2022causal}. We shall adapt the UCB approach to our context, exploiting properties of our graph identification scheme. Another standard assumption in causal bandits is that of a static causal model. Both \cite{yan2024robust} and \cite{alami2022non} consider the non-stationary case, but assume prior knowledge of the initial causal graph. We shall provide a method by which to keep the causal structure up-to-date without prior graph knowledge. A comparison between prior works and our approach is provided in Table \ref{tab:related}.

\subsection{Approach and Contributions} \label{subsec:AC}
We note that separating graph identification from optimized interventions for MABs is inefficient as these two goals are coupled in causal MABs.  Another feature not typically considered in the causal graph identification or causal MAB literature is the distinguishing between false positive and false negative errors in graph identification -- we shall actively consider this feature. A false positive means that the graph identification method outputs an edge that is not present in the true graph, while a false negative means that the graph identification method outputs no edge, where in fact, the edge is present in the true DAG. In this paper, we propose the \emph{Causal Sub-graph Learning with Upper Confidence Bound (CSL-UCB)} scheme, without assumed knowledge of either the causal graph topology or the interventional distributions.  Our algorithm seeks to minimize false negative errors in graph identification as they have a larger impact on optimal intervention design. We quantify the uncertainty in our graph identification coupled with that stemming from decision making. The learning of sub-graphs (versus the entire graph) strongly reduces complexity and provides a natural strategy for change detection in non-stationary environments. The main contributions of this paper are:
\begin{enumerate}
\item We propose a sub-graph learning approach to learn the causal structure, which has strongly improved computational complexity. Numerical results suggest that it has a much lower sample complexity as well. A novel mutual information regularization is proposed which reduces false negative errors. It is shown that false negatives have a more critical impact on overall reward optimization in causal bandit problems.
\item An uncertainty bound tailored to the causal bandit framework and our graph identification method is derived and used to drive an upper confidence bound strategy for arm selection. 
\item A sub-graph change detection mechanism with local updates is proposed enabling the consideration of non-stationary causal structure.
\item Numerical comparisons show that the proposed scheme identifies the optimal intervention much faster than standard MAB schemes by exploiting causal structure. Moreover, compared to strategies that only focus on causal graph identification, the proposed scheme needs much fewer samples than other methods and learns the causal structure with a limited loss with respect to the long-term reward.
\end{enumerate}

The rest of the paper is organized as follows. We introduce the system model in Section \ref{sec:SM}. The sub-graph learning scheme is proposed in Section \ref{sec:CSL}, with an analysis of the two types of error. Uncertainty quantification is provided in Section \ref{sec:ISU}, which enables the balance between exploration and exploitation. The sub-graph change detection mechanism is proposed in Section \ref{sec:CSCD}. Finally, the paper is concluded in Section \ref{sec:NR} and all the proofs are provided in the Appendix. 

\section{System Model} \label{sec:SM}
\subsection{The Causal Graphical Model with Soft Intervention} \label{subsec:CGM}
We assume the causal effects can be captured by a DAG with structure $(\mathcal{V}, \mathcal{B})$, where $\mathcal{V} = \{1,\dots,N\}$ is the set of $N$ nodes and $\mathcal{B}$ is the set of directed edges. Moreover, the edge-weight matrix $\bm{B} \in \reals^{N \times N}$ captures the strength of causal effects, where the $(i,j)$-th entry represents the weight of the edge $i \rightarrow j$. To describe observations under intervention, consider node-wise intervention, defined as
\begin{equation}
\bm{a} = \trans{(a_1, \dots, a_N)} \in \{0,1\}^N,
\end{equation}
where $a_i$ represents whether node $i$ is intervened ($a_i = 1$) or not ($a_i = 0$). Specifically, instead of hard interventions, we consider soft interventions, which do not necessarily cut off causal relationships between the intervened node and its parents, but change the in-coming edges to the node. To formalize the model for soft intervention, we denote the interventional edge-weight matrix by $\bm{B}' \in \reals^{N \times N}$, such that the post-intervention weight matrix $\bm{B}_{\bm{a}}$ can be constructed as
\begin{equation} \label{eq:constr_Ba}
\left[\bm{B}_{\bm{a}}\right]_i = \mathbb{I}(a_i=1) \bm{B}'_i + \mathbb{I}(a_i=0) \bm{B}_i,
\end{equation}
where $\mathbb{I}(\cdot)$ is the indicator function and $[\cdot]_i$ represents the $i$-th column of a matrix such that $\bm{B}_i, \bm{B}'_i \in \reals^{N \times 1}$. The $i$-th column of the post-interventional weight matrix determines the set of parents of node $i$ and how these parents causally influence node $i$.

As a result of intervention, the vector of stochastic values associated with the nodes is represented by $\bm{x} \in \reals^N$. The causal relationship among nodes is described by a linear structural equation model (LinSEM),
\begin{equation} \label{eq:LinSEM}
\bm{x} = \trans{(\bm{B}_{\bm{a}})} \bm{x} + \bm{\epsilon},
\end{equation}
where $\bm{\epsilon}$ is a vector of Gaussian exogenous/noise variables. We assume $\bm{\epsilon}$ contains independent elements with known means $\bm{\nu}$ and unknown variances $\bm{\sigma}^2$. Further, we denote $\tilde{\bm{\epsilon}} \doteq \bm{\epsilon} - \bm{\nu}$ such that $\tilde{\bm{\epsilon}} \sim \Gsn(\bm{0}, \mathrm{diag}(\bm{\sigma}^2))$.

We denote the sets of parents and ancestors of node $i$ by $\mathcal{P}_i(a_i)$ and $\mathcal{A}_i(a_i)$, the estimated set of parents by $\widehat{\mathcal{P}}_i(a_i)$. The set difference of the estimated and true parent sets is denoted by $\widehat{\mathcal{P}}_i \backslash \mathcal{P}_i(a_i)$. Taking node $3$ in Fig. \ref{fig:causal_graph} as an example, the parent and ancestor sets are $\mathcal{P}_3(0) = \{2\}$, $\mathcal{A}_3(0) = \{1, 2\}$. If the estimated parent set is $\widehat{\mathcal{P}}_3(0) = \{2, 4\}$, then we have $\widehat{\mathcal{P}}_3 \backslash \mathcal{P}_3(0) = \{4\}$.

\subsection{The Causal Bandit Model} \label{subsec:CBM}
\begin{figure}[!t]
\centering
\includegraphics[width=0.9\linewidth]{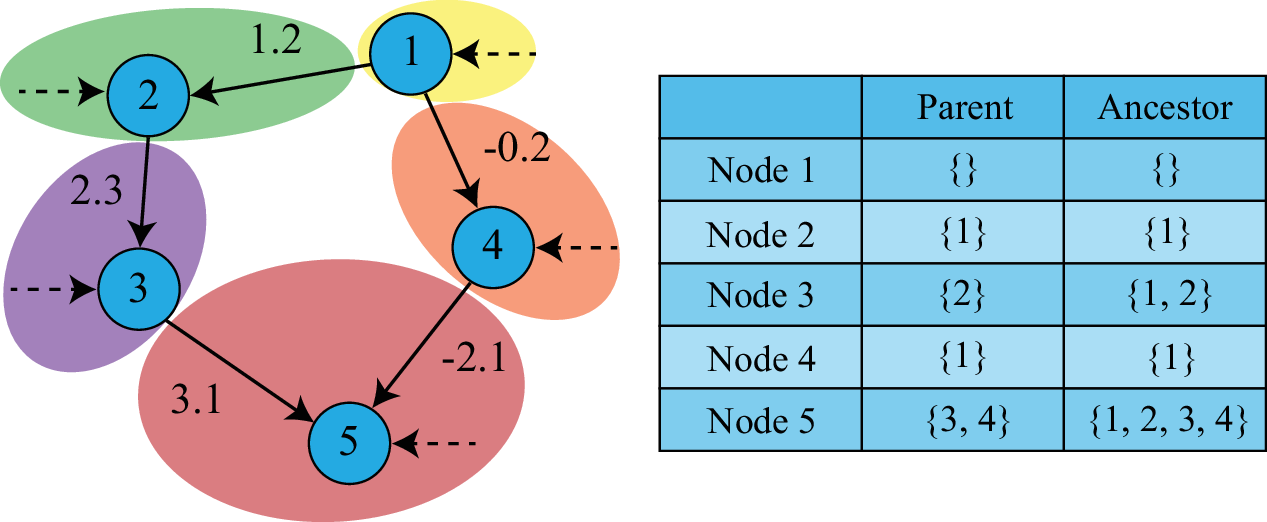}
\caption{A causal graph with $N = 5$ and node $5$ as the reward node. Each sub-graph is marked by a colored region. The sets of parents and ancestors are provided in the inline table. }
\label{fig:causal_graph}
\end{figure}

In the MAB framework, an agent performs a sequence of actions in order to maximize cumulative reward over a finite horizon $T$. With the causal graphical model, we consider node $N$ as the reward node, which generates a stochastic reward in each time step. An example of such a causal graph is given in Fig. \ref{fig:causal_graph}, where the value of node $5$ is considered the reward signal and the effect of exogenous variables are represented by dashed arrows.

To compute the expected reward under intervention $\bm{a}$, we recognize that in LinSEMs, there exists a causal flow between every ancestor-descendant pair. Thus each variable $x_i$ can be decomposed as a linear combination of exogenous variables in $\bm{\epsilon}$, weighted by the strength of the causal flow. Define the post-intervention flow-weight matrix as 
\begin{equation}
\bm{C}_{\bm{a}} \doteq \left(\bm{I} - \bm{B}_{\bm{a}} \right)^{-1}, 
\end{equation}
where the $(i,j)$-th entry represents the net flow weight from node $i$ to $j$. In this way, we can rewrite \eqref{eq:LinSEM} as
\begin{equation}
\bm{x} = \invtrans{\left(\bm{I} - \bm{B}_{\bm{a}} \right)} \bm{\epsilon} = \trans{\left(\bm{C}_{\bm{a}}\right)} \bm{\epsilon},
\end{equation}
where $\bm{I}$ denotes the identity matrix. The expectation of $\bm{x}$ under intervention $\bm{a}$ is formulated as
\begin{equation} \label{eq:mu_a}
\bm{\mu}_{\bm{a}} \doteq \mathbb{E}\left[\invtrans{\left(\bm{I} - \bm{B}_{\bm{a}} \right)} \bm{\epsilon} \right] = \invtrans{\left(\bm{I} - \bm{B}_{\bm{a}} \right)} \bm{\nu}.
\end{equation}
{Thus, given the post-intervention weight matrices, we are able to find the intervention that maximizes the expected reward, which is defines the optimal intervention,}
\begin{equation}
\bm{a}^* \doteq \argmax_{\bm{a}} \ \left[\bm{\mu}_{\bm{a}} \right]_N.
\end{equation}

In each step, the decision-making agent selects an intervention $\bm{a}^t$, observes $\bm{x}^t$ and collects a reward $x_N^t$. The randomness of the observations comes from the exogenous variables, which are independent of the intervention. The objective is to maximize the expected cumulative reward, or equivalently, minimize the expected cumulative regret, defined as
\begin{equation}
R_T \doteq \E \left[\sum_{t = 1}^T \left(\left[\bm{\mu}_{\bm{a}^*} \right]_N - \left[\bm{\mu}_{\bm{a}}^t \right]_N \right) \right].
\end{equation}

We underscore that all key derivations and theoretical results in the sequel assume the signal model in this section.

\section{Causal Sub-graph Learning} \label{sec:CSL}
Causal bandit problems are different from ordinary MAB problems, due to the existence of an underlying causal structure, which is assumed to be unknown in our setting. Since the expected reward is a function of the post-intervention weight matrix $\bm{B}_{\bm{a}}$, understanding the causal structure enables the agent to better select the intervention afterwards. Over a finite horizon, the agent has to balance between gaining causal knowledge (exploration) and maximizing immediate reward (exploitation), which is a common dilemma for sequential decision-making problems. In this section, we focus on identifying the causal structure, in a sub-graph learning manner. Based on that, the problem of balancing exploration and exploitation is investigated in the following section.

\subsection{Sub-graph Decomposition}
Causal graph identification is equivalent to figuring out causal relationships between every pair of nodes in the graph, represented by directed edges. Given observation up to step $t$, $\bm{X}^{1:t} \doteq (\bm{x}^1, \dots, \bm{x}^t)$ of dimension $N \times t$, we can evaluate every possible graph structure by its ability to fit the data. However, finding the best graph structure may not be feasible in practice because the number of possible DAGs grows super-exponentially in the number of nodes. For example, the number of DAGs for $2$, $4$, $10$ nodes are $3$, $543$ and $4.2 \times 10^{18}$, respectively \cite{peters2017elements}. The growing search space issue becomes even more serious for causal bandits. Since the effect of an intervention is characterized by the post-interventional graph $\bm{B}_{\bm{a}}$ and there are $2^N$ possible interventions, the agent has to identify $2^N$ different graphs with only $t$ observations.

To improve sample efficiency, one feature of the causal bandit model is noteworthy. Even though there are $2^N$ post-intervention distributions characterized by $\bm{B}_{\bm{a}}$, those post-intervention weight matrices are composed of columns of $\bm{B}$ and $\bm{B}'$ \cite{varici2023causal}. Therefore, instead of identifying $2^N$ causal graphs induced by $\bm{B}_{\bm{a}}$, we can identify causal relationships induced by the columns of $\bm{B}$ and $\bm{B}'$. Specifically, $\bm{B}_i$ and $\bm{B}'_i$ encode causal weights between node $i$ and its parents. The following proposition ensures that causal graphs can always be uniquely decomposed into sub-graphs of our interest.\footnote{The sub-graphs we consider here are not vertex-induced sub-graphs. }

\begin{prop} \label{prop:decompose}
Any DAG can always be uniquely decomposed into sub-graphs such that each sub-graph consists of a node and all in-coming edges to the node.
\end{prop}
\begin{IEEEproof}
First, we prove the existence of such decomposition by construction. For any complete DAG with a set of nodes $\mathcal{V} = \{1,\dots,N\}$ and a set of edges $\mathcal{B}$, we can always decompose it into a group of sub-graphs $\mathcal{G}_i, i \in \{1,\dots,N\}$, such that 
\begin{equation} \label{eq:def_subgraph}
\mathcal{V}_i = \{i\}, \quad \mathcal{B}_i = \{(k, i): \forall k \in \mathcal{P}_i \}.
\end{equation}
Moreover, notice that the decomposition is unique: sub-graph $\mathcal{G}_i$ contains and only contains node $i$ and all of its in-coming edges, which completes the proof.
\end{IEEEproof}

As illustrated in Fig. \ref{fig:causal_graph}, each sub-graph resides in one of the colored regions. For example, the sub-graph in the red region consists of node $5$, directed edges $(3, 5)$ and $(4, 5)$. Learning sub-graphs has exponentially improved data efficiency. Consider an example where interventions are selected in turns up to step $t$, such that each complete graph is associated with $t/2^N$ data samples, as there are $2^N$ complete graphs. Each sub-graph, instead, is associated with $t/2$ data samples, as each node can only be in one of two possible modes: under intervention, or not.

For learning causal sub-graphs, we employ the \emph{principle of independent mechanisms}, which states that the causes and the mechanism producing the effect are independent \cite{peters2017elements}. Based on that, testing the independence of residuals has been investigated for causal discovery, in the context of additive noise models \cite{mooij2016distinguishing,ghassami2018multi,wu2020information}. Herein, we consider minimum mean-square error (MMSE) estimation, where the estimator and residual are defined as
\begin{equation}
\hat{x}_i^t(a_i) \doteq \E \left[x_i \middle| \bm{X}^{1:t}_{\widehat{\mathcal{P}}_i(a_i)} \right], \quad r_i^t(a_i) \doteq \hat{x}_i^t(a_i) - x_i.
\end{equation}
With the estimated mechanism producing $x_i$, the principle of independent mechanisms implies that the residual should be independent from values of the parent nodes,
\begin{equation}
r_i^t(a_i) \perp x_j, \quad \forall j \in \mathcal{P}_i(a_i).
\end{equation}
Thus, the dependence between $r_i^t$ and $x_j$ supports the hypothesis that node $j$ is not a parent of node $i$ and thus the edge $j \rightarrow i$ does not exist. By evaluating all potential in-coming edges to a node, the corresponding sub-graph can be estimated. 

\subsection{Edge-weighted Mutual Information} \label{subsec:EWMI}
As a general measure of dependence, mutual information (MI), denoted by $I(\cdot)$, is widely utilized for causal discovery (see \textit{e.g.} \cite{janzing2010causal,sun2016mutual}). However, since the sub-graphs are treated separately, pure MI-based sub-graph learning with limited data suffers \emph{significantly from estimation error}. In this subsection, we explain this issue in detail and propose the \emph{edge-weighted mutual information} measure as a solution.

We take a closer look at the mutual information of interest. In the following derivation, we focus on node $i$ under intervention $\bm{a}$ and a potential parent node $j$, with observation up to step $t$. The true causal relationship is represented as
\begin{equation} \label{eq:xi_decomp}
x_i = \sum_{k \in \mathcal{P}_i(a_i)} \left[\bm{B}_{\bm{a}}\right]_{ki} x_k + \epsilon_i = \sum_{k \in \mathcal{A}_i(a_i)} \left[\bm{C}_{\bm{a}}\right]_{ki} \epsilon_k + \epsilon_i.
\end{equation}
For example, we have $x_3 = 2.76 \epsilon_1 + 2.3 \epsilon_2 + \epsilon_3$ for node $3$ in Fig. \ref{fig:causal_graph}. With the estimated edge-weight matrix denoted by $\widehat{\bm{B}}_{\bm{a}}^t$, the estimated flow-weight matrix is defined as
\begin{equation}
\widehat{\bm{C}}_{\bm{a}}^t \doteq \bigl(\bm{I} - \widehat{\bm{B}}_{\bm{a}}^t \bigr)^{-1}.
\end{equation}
Then the estimated node value can be decomposed using flow weights as
\begin{equation} \label{eq:xi_hat_decomp}
\hat{x}_i^t = \sum_{k \in \mathcal{A}_i(a_i)} \bigl[\widehat{\bm{C}}_{\bm{a}}^t\bigr]_{ki} \epsilon_k + \sum_{l \not\in \mathcal{A}_i(a_i)} \bigl[\widehat{\bm{C}}_{\bm{a}}^t\bigr]_{li} \epsilon_l + \nu_i.
\end{equation}
For example, the estimate of $x_3$ can be decomposed according to the ancestor set of node $3$, as presented in Fig. \ref{fig:causal_graph}. Since we do not assume any knowledge of the causal structure, all other nodes are considered as potential parents of node $i$. Based on \eqref{eq:xi_decomp} and \eqref{eq:xi_hat_decomp}, the residual can be expressed as
\begin{multline} \label{eq:res}
r_i^t(a_i) = - \tilde{\epsilon}_i + \\
\underbrace{\sum_{k \in \mathcal{A}_i(a_i)} \bigl(\bigl[\widehat{\bm{C}}_{\bm{a}}^t\bigr]_{ki} -  \left[\bm{C}_{\bm{a}}\right]_{ki}\bigr) \epsilon_k}_{\textstyle\text{intrinsic error}} + \underbrace{\sum_{l \not\in \mathcal{A}_i(a_i)} \bigl[\widehat{\bm{C}}_{\bm{a}}^t\bigr]_{li} \epsilon_l}_{\textstyle\text{causal error}}.
\end{multline}
The second term in \eqref{eq:res} is always non-zero in the finite sample regime, thus we define this error as the \emph{intrinsic error}. The last term is defined as the \emph{causal error} because it is non-zero only when some non-ancestor nodes are considered as direct causes (otherwise the residual would not contain $\epsilon_l$). For a specific non-ancestor node $j$, which is a source of the causal error, we decompose its value as
\begin{equation}
x_j = \sum_{k \in \mathcal{A}_i(a_i)} \left[\bm{C}_{\bm{a}}\right]_{kj} \epsilon_k + \sum_{l \not\in \mathcal{A}_i(a_i)} \left[\bm{C}_{\bm{a}}\right]_{lj} \epsilon_l.
\end{equation}
Now that we have both $r_i^t(a_i)$ and $x_j$ for mutual information evaluation, the following proposition provides an upper bound where the impact of intrinsic and causal errors are separated.

\begin{prop} \label{prop:I_decomp}
With independent exogenous variables, the mutual information between the residual and a declared parent can be upper bounded as
\begin{align} \label{eq:MI_decomp}
I(r_i^t(a_i);x_j) \leq \underbrace{I\Bigl(\sum_{l \not\in \mathcal{A}_i(a_i)} \bigl[\widehat{\bm{C}}_{\bm{a}}^t\bigr]_{li} \epsilon_l - \tilde{\epsilon}_i; \sum_{l \not\in \mathcal{A}_i(a_i)} \left[\bm{C}_{\bm{a}}\right]_{lj} \epsilon_l \Bigr)}_{\textstyle\text{causal error induced MI}} & \nonumber \\
+ \underbrace{I\Bigl(\sum_{k \in \mathcal{A}_i(a_i)} \bigl(\bigl[\widehat{\bm{C}}_{\bm{a}}^t\bigr]_{ki} -  \left[\bm{C}_{\bm{a}}\right]_{ki}\bigr) \epsilon_k; \sum_{k \in \mathcal{A}_i(a_i)} \left[\bm{C}_{\bm{a}}\right]_{kj} \epsilon_k \Bigr)}_{\textstyle\text{intrinsic error induced MI}}& .
\end{align}
\end{prop}
\begin{IEEEproof}
See Appendix \ref{proof:prop_I_decomp}.
\end{IEEEproof}

The proof relies on the data processing inequality \cite{cover1999elements} and the mutual independence of the exogenous variables. We define the first and second terms on the right-hand side of \eqref{eq:MI_decomp} as the mutual information induced by the causal and intrinsic errors, respectively. Due to the existence of the intrinsic error, we may not be able to detect the causal error and thus misjudge causal directions. To illustrate this point, consider the \emph{causal error induced mutual information}. Denote the linear combination of exogenous variables from non-ancestors by
\begin{equation} \label{eq:phi_x}
\theta(x_j,i) \doteq \sum_{l \not\in \mathcal{A}_i(a_i)} \left[\bm{C}_{\bm{a}}\right]_{lj} \epsilon_l,
\end{equation}
which is Gaussian, with mean $\mu_{\theta x}$ and variance $\sigma^2_{\theta x}$. For example, since the ancestor set of node $3$ in Fig. \ref{fig:causal_graph} is $\{1, 2\}$, $\theta(x_5,3)$ is a linear combination of $\epsilon_3$, $\epsilon_4$ and $\epsilon_5$. As we have a causal error, there is an assumed causal edge that does not exist in the true graph. This erroneous causal edge, $j \to i$ (from non-ancestors) is used in the estimation, and as a result, induces a causal error in the residual, weighted by the estimated parameter. To focus on the effect caused by the erroneously labeled causal edge $j \rightarrow i$, we denote the causal error in the residual induced by other nodes by
\begin{equation}
\theta(r_i^t(a_i), \backslash j) \doteq \sum_{l \not\in \mathcal{A}_i(a_i)} \bigl[\widehat{\bm{C}}_{\bm{a}}^t\bigr]_{li} \epsilon_l - \tilde{\epsilon}_i - \bigl[\widehat{\bm{B}}_{\bm{a}}^t\bigr]_{ji} \theta(x_j,i),
\end{equation}
which is Gaussian as well, with mean $\mu_{\theta r}$ and variance $\sigma^2_{\theta r}$, while $\backslash j$ denotes the set of other non-ancestors, $\backslash j \doteq \{l \in \{1,\dots,N\} | l \notin \mathcal{A}_i(a_i), l \neq j \}$. In this way, we can rewrite and bound the causal error induced mutual information as
\begin{align}
&I\Bigl(\sum_{l \not\in \mathcal{A}_i(a_i)} \bigl[\widehat{\bm{C}}_{\bm{a}}^t\bigr]_{li} \epsilon_l - \tilde{\epsilon}_i; \sum_{l \not\in \mathcal{A}_i(a_i)} \left[\bm{C}_{\bm{a}}\right]_{lj} \epsilon_l \Bigr) \nonumber \\
&\quad = I\Bigl(\bigl[\widehat{\bm{B}}_{\bm{a}}^t\bigr]_{ji} \theta(x_j,i) + \theta(r_i^t(a_i), \backslash j); \theta(x_j,i) \Bigr) \\
&\quad = h\Bigl(\bigl[\widehat{\bm{B}}_{\bm{a}}^t\bigr]_{ji} \theta(x_j,i) + \theta(r_i^t(a_i), \backslash j)\Bigr) + h\Bigl(\theta(x_j,i)\Bigr) \nonumber \\ 
&\hspace{108pt} - h\Bigl(\theta(r_i^t(a_i), \backslash j), \theta(x_j,i) \Bigr) \\
&\quad \leq \log\Bigl(\bigl(|\bigl[\widehat{\bm{B}}_{\bm{a}}^t\bigr]_{ji}| \sigma_{\theta x} + \sigma_{\theta r}\bigr) \sqrt{2\pi e}\Bigr) + \log\Bigl(\sigma_{\theta x} \sqrt{2\pi e} \Bigr) \nonumber \\
&\hspace{62pt} - \left[\log(2\pi e) + \log\Bigl(\sigma_{\theta x} \sigma_{\theta r} \sqrt{1 - \rho_{\theta}^2}\Bigr) \right] \label{eq:MI_indep_bound} \\
&\quad = \log \Bigl(1 + \frac{\sigma_{\theta x}}{\sigma_{\theta r}} |\bigl[\widehat{\bm{B}}_{\bm{a}}^t\bigr]_{ji}| \Bigr) - \frac{1}{2} \log(1 - \rho_{\theta}^2),
\end{align}
where $\rho_{\theta}$ stands for the correlation
\begin{equation}
\rho_{\theta} = \frac{\E[(\theta(x_j,i) - \mu_{\theta x}) (\theta(r_i^t(a_i), \backslash j) - \mu_{\theta r})]}{\sigma_{\theta x} \sigma_{\theta r}},
\end{equation}
and $h(\cdot)$ denotes the differential entropy. Note that the inequality \eqref{eq:MI_indep_bound} comes from the Gaussianity of key random variables and the fact that the variance is maximized when the variables are perfectly correlated ($\rho_{\theta} = 1$).

We observe that the causal error induced mutual information becomes small when the estimated edge weight is small. To make matters worse, the intrinsic error induced mutual information is normally not negligible in the limited data regime. Thus, it is challenging to detect the causal error and reject a non-causal edge when the estimated weight is small. To alleviate this issue, we propose an \emph{edge-weighted mutual information} measure, defined as
\begin{equation}
I_{\mathrm{w}}(r_i^t(a_i); x_j) \doteq I(r_i^t(a_i); x_j) - \log |\bigl[\widehat{\bm{B}}_{\bm{a}}^t\bigr]_{ji}|.
\end{equation}
A large $I_\mathrm{w}$ indicates that the corresponding edge is non-causal and thus should be rejected. The remaining problem is to determine the threshold for rejecting edges. 

In general, any residual based test for causal identification needs to balance between false positive and false negative errors \cite{peters2017elements}, by selecting an appropriate testing threshold. However, in terms of maximizing cumulative reward, the two types of error contribute differently to the ultimate error of reward estimation. To be more specific, we argue that rejecting an actual edge (FN error) is much worse than accepting a nonexistent edge (FP error). Formally, FN errors exist in the estimated causal graph (observational or interventional) if 
\begin{equation} \label{eq:crit_fail}
\exists (i, j): \quad (i, j) \in \mathcal{B}, \quad (i, j) \notin \widehat{\mathcal{B}},
\end{equation}
where $\widehat{\mathcal{B}}$ denotes the estimated edge set. In the following subsections, we analyze the impact of FN and FP errors and propose an adaptive edge-rejecting strategy to minimize the FN error rate.

\subsection{Analysis of Graph Errors} \label{subsec:AGE}
As discussed in Section \ref{subsec:CBM}, estimated rewards under different interventions are computed based on the estimated weight matrices. Thus, we can evaluate the impact of the two types of error by their effects on weight matrix estimation. Focusing on the sub-graph of node $i$ under intervention $a_i$, the following proposition gives the estimated edge weights in expectation, under the two types of errors.

\begin{prop} \label{prop:graph_error} 
We assume MMSE estimation and determine the effect of \textbf{only} \emph{false positive} or \textbf{only} \emph{false negative} errors. For the presence of only \emph{false positive} errors, the estimated edge weights satisfy
\begin{equation}
\E\bigl[\bigl[\widehat{\bm{B}}_{\bm{a}}^t\bigr]_{i,\widehat{\mathcal{P}}_i(a_i)} \big| \bm{X}^{1:t}_{\widehat{\mathcal{P}}_i(a_i)}(a_i)\bigr] = \left[\bm{B}_{\bm{a}}\right]_{i,\widehat{\mathcal{P}}_i(a_i)}.
\end{equation}
With only \emph{false negative} errors, the estimated edge weights satisfy
\begin{align}
&\E\bigl[\bigl[\widehat{\bm{B}}_{\bm{a}}^t\bigr]_{i,\widehat{\mathcal{P}}_i(a_i)} \big|\bm{X}^{1:t}_{\widehat{\mathcal{P}}_i(a_i)}(a_i)\bigr] = \left[\bm{B}_{\bm{a}}\right]_{i,\widehat{\mathcal{P}}_i(a_i)} + \nonumber \\
&\hspace{16pt} \Bigl[\bm{X}^{1:t}_{\widehat{\mathcal{P}}_i(a_i)}(a_i) \trans{\bm{X}^{1:t}_{\widehat{\mathcal{P}}_i(a_i)}(a_i)}\Bigr]^{-1} \E\Bigl[\bm{X}^{1:t}_{\widehat{\mathcal{P}}_i(a_i)}(a_i) \cdot \nonumber \\
&\hspace{32pt} \trans{\bm{X}^{1:t}_{\mathcal{P}_i \backslash \widehat{\mathcal{P}}_i(a_i)}(a_i)} \Big| \bm{X}^{1:t}_{\widehat{\mathcal{P}}_i(a_i)}(a_i)\Bigr] \left[\bm{B}_{\bm{a}}\right]_{i,\mathcal{P}_i \backslash \widehat{\mathcal{P}}_i(a_i)}. \label{eq:Ba_error_cf}
\end{align}
\end{prop}
\begin{IEEEproof}
See Appendix \ref{proof:prop_graph_error}.
\end{IEEEproof}
The proof is based on the properties of MMSE estimation. Note that $\bm{X}^{1:t}$ and $\bm{\epsilon}^{1:t}$ have $a_i$ as an argument, because only the values associated with the intervention $a_i$ are considered for learning the causal structure under intervention $a_i$.

Observe that with only FP errors, the estimated edge weights are asymptotically unbiased. Although the estimated weights for any non-existent parents in $\widehat{\mathcal{P}}_i(a_i)$ is non-zero for a finite number of samples, the bias converges to zero as more samples are collected. In contrast, with the existence of FN errors, the estimated weights for both correctly identified and erroneously rejected parents are biased, resulting in inaccurate weights even with an infinite amount of data. Note that the bias is a function of the correlation between identified parents and rejected parents.

\subsection{Adaptive Edge Rejection}
Since estimated rewards are inaccurate when FN errors occur, we should avoid rejecting an actual edge and ensure the complete graph is a DAG. Thus, the threshold for rejecting edges should be set large enough to minimize the probability of rejecting an actual edge. On the other hand, the threshold should not be too large because redundant edges can result in cycles in the complete causal graph.

To achieve this goal, we propose to adjust the threshold adaptively such that edges are rejected only when necessary. To be more specific, the edge-weighted mutual information is calculated for each potential edge and the edge with the largest $I_{\mathrm{w}}$ is rejected at a time, until the complete graph becomes a DAG. Once the set of declared edges is determined, we construct the weight matrices by MMSE estimation. Notice that in this process, although sub-graphs are learned separately with different sets of samples, whether an edge should be rejected or not is also influenced by edges in other sub-graphs.

The overall approach for causal graph identification is denoted as the \emph{Causal Sub-graph Learning (CSL)} scheme, and the pseudo-code is provided in Algorithm \ref{alg:CSL}. Note that although the observational weight matrix $\bm{B}$ is used as the target in the pseudo-code, the same procedure will be used for learning the interventional weight matrix $\bm{B}'$ as well.

\begin{algorithm}[!t]
\begin{algorithmic}[1]
\caption{The Causal Sub-graph Learning Scheme}
\label{alg:CSL}
\REQUIRE{The set of nodes $\mathcal{V}$ and node values $\bm{X}^{1:t}$. }
\STATE Initialize the edge set to include all possible directed edges: $\widehat{\mathcal{B}} = \{(i,j): \forall i, j \in \mathcal{V}\}$.
\WHILE{$(\mathcal{V}, \widehat{\mathcal{B}})$ is not a DAG}
\STATE Compute edge weights $\widehat{\bm{B}}_{ij}$ and residuals, $\forall (i,j) \in \widehat{\mathcal{B}}$, by MMSE estimation, with observed values $\bm{X}^{1:t}$.
\STATE Compute edge-weighted MI $I_{\mathrm{w}}(r_i^t(0); x_j)$, $\forall (i,j) \in \widehat{\mathcal{B}}$, with $\bm{X}^{1:t}$ and empirical MI.
\STATE Find $(i,j) = \argmax_{i,j} I_{\mathrm{w}}(r_i^t(0); x_j)$, remove the edge $(i,j)$ from $\widehat{\mathcal{B}}$.
\ENDWHILE
\STATE Compute $\widehat{\bm{B}}_{ij}$, $\forall (i,j) \in \widehat{\mathcal{B}}$, by MMSE estimation, with $\bm{X}^{1:t}$. For $(i,j) \notin \widehat{\mathcal{B}}$, set $\widehat{\bm{B}}_{ij} = 0$.
\RETURN{Estimated weight matrix $\widehat{\bm{B}}$. }
\end{algorithmic}
\end{algorithm}

Lastly, we provide the time and space complexity of the proposed algorithm. Starting with a set of all possible edges of size $N^2$, the algorithm rejects an edge in each step until the graph becomes a DAG, and thus the number of steps is less than $N^2$. Each step selects an edge with the largest edge-weighted MI among the remaining edges, with time complexity $\mathcal{O}(N^2)$. After the rejection of an edge, estimation of the weights and residuals takes $\mathcal{O}(N^2(N + t))$ time while estimation of the MI takes $\mathcal{O}(Nt\log t)$ time \cite{vollmer2018complexity}. Besides, checking whether the graph is a DAG by topological sorting takes $\mathcal{O}(N^2)$ time \cite{cormen2022introduction}. Finally, if $N \leq t$, the time complexity of the proposed scheme is $\mathcal{O}(N^4 t + N^3 t\log t)$. The space complexity is $\mathcal{O}(Nt)$ for storing the collected samples and a matrix of edge-weighted MI.

\section{Intervention Selection under Uncertainty} \label{sec:ISU}
Given the ability of learning causal graphs with observed data, we can estimate the reward under each possible intervention. Although selecting the intervention with largest estimated reward is optimal for the current step, exploring other interventions is necessary for achieving higher cumulative reward in the long run. It is because the intervention with largest estimated reward may not be the optimal intervention, due to the uncertainty of reward estimation. Thus, intervention selection should consider both how close the estimated rewards are to being maximal and the uncertainties in those estimates. In this section, we derive an uncertainty bound on the estimation error of the expected reward, so that the potential of an intervention can be taken into account. 

For any specific intervention $\bm{a}$, define the matrix of edge-weight errors and the vector of expected node-value errors as
\begin{equation}
\Delta\bm{B}_{\bm{a}}^t \doteq \widehat{\bm{B}}_{\bm{a}}^t - \bm{B}_{\bm{a}}, \quad \Delta\bm{\mu}_{\bm{a}}^t \doteq \hat{\bm{\mu}}_{\bm{a}}^t - \bm{\mu}_{\bm{a}},
\label{eq:est_reward}
\end{equation}
where $\hat{\bm{\mu}}_{\bm{a}}^t$ represents the estimated mean of $\bm{x}$. As analyzed in Section \ref{subsec:AGE}, the weight errors could have non-zero means, due to missing edges (false negatives). With the centered error matrix defined as
\begin{equation}
\Delta\tilde{\bm{B}}_{\bm{a}}^t \doteq \Delta\bm{B_a}^t - \E\bigl[\Delta\bm{B}_{\bm{a}}^t \big| \bm{X}^{1:t}(\bm{a})\bigr],
\end{equation}
we bound the error in the estimated reward (see \eqref{eq:est_reward}) as
\begin{align}
\left|\left[\Delta\bm{\mu}_{\bm{a}}^t\right]_N\right| &= \left|\trans{\bigl[(\bm{I}-\widehat{\bm{B}}_{\bm{a}}^t)^{-1}\bigr]}_N \trans{(\Delta\bm{B_a}^t)} \bm{\mu}_{\bm{a}}\right| \\
&\leq \left|\trans{\bigl[(\bm{I}-\widehat{\bm{B}}_{\bm{a}}^t)^{-1}\bigr]}_N \trans{\E\bigl[\Delta\bm{B}_{\bm{a}}^t \big| \bm{X}^{1:t}(\bm{a})\bigr]} \bm{\mu}_{\bm{a}}\right| \nonumber \\
&\hspace{40pt} + \left|\trans{\bigl[(\bm{I}-\widehat{\bm{B}}_{\bm{a}}^t)^{-1}\bigr]}_N \trans{\bigl(\Delta\tilde{\bm{B}}_{\bm{a}}^t\bigr)} \bm{\mu}_{\bm{a}}\right|. \label{eq:del_mu_ub}
\end{align}
Recall that we provide an expression for the non-zero elements of $\E\bigl[\Delta\bm{B}_{\bm{a}}^t \big| \bm{X}^{1:t}(\bm{a})\bigr]$ in \eqref{eq:Ba_error_cf}, which depends on the correlation between correctly declared parents and mistakenly rejected parents. However, since we are not able to know whether and where causal edges are rejected by mistake, the uncertainty bound cannot be computed at run-time.

Nevertheless, when no FN error exists, we are able to bound the uncertainty of the estimated reward. With the covariance of the $i$-th weight error vector denoted by $\bm{\Phi}_{\bm{a}}^t(i)$, we derive a concentration inequality for the estimation error, under the assumption of no FN errors.

\begin{lemma} \label{lem:moment_bound}
With MMSE estimation and the assumption of no false negative errors, the norm of the weight error vectors have upper-bounded moments for $m \geq 2$:
\begin{equation}
\E \Bigl[\left\lVert \left[\Delta\bm{B}_{\bm{a}}^t\right]_i \right\rVert_2^m \Bigr] \leq
m! \sqrt{\frac{4N}{N+2}} \left[\lambda_{\max}(\bm{\Phi}_{\bm{a}}^t(i)) \frac{N+2}{4} \right]^{m/2},
\end{equation}
where $\lambda_{\max}(\cdot)$ stands for the maximum eigenvalue of a matrix.
\end{lemma}
\begin{IEEEproof}
See Appendix \ref{proof:lem_moment_bound}.
\end{IEEEproof}
The proof is based on the eigenvalue decomposition of the weight error matrix and Jensen's inequality. Lemma \ref{lem:moment_bound} enables us to derive a concentration inequality for the maximum singular-value of the weight error matrix.

\begin{lemma} \label{lem:sv_bound}
With MMSE estimation and the assumption of no false negative error, the following inequality holds,
\begin{multline}
\P\biggl\{\sigma_{\max}(\Delta\bm{B}_{\bm{a}}^t) \geq 2\left(N^2+2N\right)^{1/4} \\
\times \sqrt{\ln\Bigl(\frac{2N}{\delta}\Bigr) \sum_{i=1}^N \lambda_{\max}(\bm{\Phi}_{\bm{a}}^t(i))} \biggr\} \leq \delta,
\end{multline}
where $\sigma_{\max}(\cdot)$ represents the maximum singular-value.
\end{lemma}
\begin{IEEEproof}
See Appendix \ref{proof:lem_sv_bound}.
\end{IEEEproof}
The proof relies on a key dilation transformation, Lemma \ref{lem:moment_bound} and the matrix Bernstein inequality \cite{tropp2012user}. Based on Lemma \ref{lem:sv_bound}, the following theorem provides a concentration inequality for the error of the estimated reward.

\begin{theorem} \label{thm:conf_bound}
Under the assumption of MMSE estimation and no false negative errors, the following inequality holds:
\begin{equation} \label{eq:conc_error_mu}
\P\biggl\{\left|\left[\Delta\bm{\mu}_{\bm{a}}^t\right]_N\right| \geq U(\bm{X}^{1:t}, \bm{a}, \delta) \biggr\} \leq \delta,
\end{equation}
where $U(\bm{X}^{1:t}, \bm{a}, \delta)$ represents the error upper-bound at confidence level $1 - \delta$,
\begin{multline} \label{eq:U}
U(\bm{X}^{1:t}, \bm{a}, \delta) = 2\left(N^2+2N\right)^{1/4} \left\lVert \bigl[(\bm{I}-\widehat{\bm{B}}_{\bm{a}}^t)^{-1}\bigr]_N \right\rVert_2 \\
\times \left\lVert \bm{\mu}_{\bm{a}} \right\rVert_2 \ \sqrt{\ln\Bigl(\frac{2N}{\delta}\Bigr) \sum_{i=1}^N \lambda_{\max}(\bm{\Phi}_{\bm{a}}^t(i))}.
\end{multline}
\end{theorem}
\begin{IEEEproof}
See Appendix \ref{proof:thm_conf_bound}.
\end{IEEEproof}
The proof is based on the Woodbury matrix identity \cite{higham2002accuracy} and Lemma \ref{lem:sv_bound}. Since the uncertainty of reward estimation cannot be measured in the presence of FN errors, we employ \eqref{eq:conc_error_mu} as an approximation for the general case. The bound on uncertainty enables us to balance between exploration and exploitation, so that the cumulative regret is minimized in the long run. In the presence of FN errors, although the uncertainty bound may not hold exactly, the observed low cumulative regret in Section \ref{subsec:PSCB} suggests, at least numerically, that the uncertainty bound has utility in our proposed algorithm.

Note that the uncertainty bounds are computed based on the estimation of edge-weight matrices. Initially, to ensure there exist samples for learning every sub-graph, we select interventions randomly. This phase is thus our \emph{exploration} start.  Afterwards, we select the intervention that achieves the best balance between exploitation (large estimated reward) and exploration (high uncertainty) in each time step via:
\begin{equation} \label{eq:select_a}
\bm{a}^{t+1} = \argmax_{\bm{a}} \left\{\trans{\bigl[(\bm{I} - \widehat{\bm{B}}_{\bm{a}}^t)^{-1}\bigr]}_N \ \bm{\nu} + \alpha U(\bm{X}^{1:t}, \bm{a}, \delta) \right\},
\end{equation}
where $\alpha$ is a design parameter that controls the exploration level. Combining CSL with uncertainty evaluation, the overall approach is denoted as the \emph{Causal Sub-graph Learning with Upper Confidence Bound (CSL-UCB)} scheme, whose pseudo-code is provided in Algorithm \ref{alg:CSL-UCB}.

\begin{algorithm}[!t]
\begin{algorithmic}[1]
\caption{The CSL-UCB Scheme}
\label{alg:CSL-UCB}
\REQUIRE{The set of nodes $\mathcal{V}$, length of exploring start $T_{\mathrm{es}}$, uncertainty level $\delta$ and exploration parameter $\alpha$. }
\HEADER{Exploring Start Phase}
\FOR{$t = 1:T_{\mathrm{es}}$}
\STATE Randomly select an intervention $\bm{a}^t$.
\STATE Observe node values $\bm{x}^t$ and collect a reward of $x_N^t$.
\ENDFOR
\ENDHEADER
\HEADER{Balanced Exploitation \& Exploration Phase}
\FOR{$t = T_{\mathrm{es}}+1:T$}
\STATE Estimate the observational and interventional causal graphs using Algorithm \ref{alg:CSL}.
\STATE For each intervention, construct the post-intervention weight matrix by \eqref{eq:constr_Ba} and evaluate the uncertainty bound according to \eqref{eq:U}.
\STATE Select an intervention $\bm{a}^t$ according to \eqref{eq:select_a}.
\STATE Observe node values $\bm{x}^t$ and collect a reward of $x_N^t$.
\ENDFOR
\ENDHEADER
\end{algorithmic}
\end{algorithm}

\section{Causal Sub-graph Change Detection} \label{sec:CSCD}
So far, we have investigated the causal bandit problem, assuming that the underlying causal structure is invariant. However, it could be non-stationary in real-world scenarios, such that the post-intervention weight matrix $\bm{B}_{\bm{a}}$ varies over time. Although the proposed CSL-UCB algorithm keeps updating the causal structure, it should only utilize data generated by the current causal model. In this section, we develop a causal sub-graph \emph{change detector} for both weight and topology changes, which enables the proposed scheme to maintain a high cumulative reward even when the causal structure is non-stationary.

Consider a piece-wise stationary model where the causal structure remains the same between two consecutive change points. Without loss of generality, we focus on detecting the first change-point in time, denoted by $t_c$. To determine whether a change occurs at step $s$, a straightforward approach compares the distribution of observed values during $[1,s]$ with the distribution during $(s,t]$. If the distributions differ significantly, $s$ is claimed to be a change point. However, the distribution shift of $\bm{x}$ only indicates the existence of a change but not the source of the change. To understand this, recall the data generating process, 
\begin{equation}
x_i = \sum_{k \in \mathcal{P}_i(a_i)} \left[\bm{B}_{\bm{a}}\right]_{ki} x_k + \epsilon_i,
\end{equation}
where the distribution shift of $x_i$ could be due to the change of causal mechanism $\left[\bm{B}_{\bm{a}}\right]_i$ or/and the distribution shifts of its parents $x_k$. As the source of change is not identifiable, the whole graph has to be re-learned.

Since causal mechanisms are modular and independent \cite{peters2017elements}, the sparse mechanism shift hypothesis \cite{scholkopf2021toward} suggests that changes in the data distribution generally are caused by changes in only a subset of causal mechanisms. In this case, it is sufficient to only re-learn the changed part. We rely on the whitening transformation of $\bm{x}$ to identify the set of changed mechanisms,
\begin{align}
\bm{y} &\doteq \trans{(\bm{I} - \widehat{\bm{B}}_{\bm{a}}^t)} \bm{x} = \trans{(\bm{I} - \bm{B}_{\bm{a}})} \bm{x} - \trans{(\Delta\bm{B}_{\bm{a}}^t)} \bm{x} \\
&= \bm{\epsilon} - \trans{(\Delta\bm{B}_{\bm{a}}^t)} \bm{x}. \label{eq:whiten_x}
\end{align}
According to \eqref{eq:whiten_x}, $y_i$ and $\epsilon_i$ differ in distribution if and only if $\bigl[\Delta\bm{B}_{\bm{a}}^t\bigr]_i \neq \bm{0}$, in which case the estimated sub-graph model fails to explain the observed data well. There are two main possible reasons for a large discrepancy between distributions: causal structure change and FN errors in estimation. Thus, besides detecting causal structure changes, checking the consistency between $\bm{y}$ and $\bm{\epsilon}$ also enables us to discover FN errors.

When the sub-graph associated with node $i$ changes, the whitened variable $y_i^{\tau}$ would differ in distribution from $\epsilon_i$. Thus, to determine whether there is a change point, we formulate a hypothesis testing problem,
\begin{align}
\mathcal{H}_0: \forall \tau \in [1, t], \ &y_i^{\tau} \iid \Gsn(\nu_i, \sigma_i^2), \\
\mathcal{H}_1: \exists t_c \in (1, t), \ &y_i^{\tau} \iid \begin{cases}
\Gsn(\nu_i, \sigma_i^2), & \tau \in [1, t_c] \\
\Gsn(\nu_i', \sigma_i'^2), & \tau \in (t_c, t] \\
\end{cases},
\end{align}
where $\nu_i$, $\sigma_i^2$ are the mean and variance of $\epsilon_i$, while $\nu_i'$, $\sigma_i'^2$ are the mean and variance of $y_i^{\tau}$, $\tau \in (t_c, t]$.

For testing the null hypothesis of no change-point versus the alternative hypothesis of a single change-point, we employ the Generalized Likelihood Ratio (GLR) statistic \cite{lai2010sequential},
\begin{align}
\Psi_i(s) \doteq \sum_{\tau = 1}^s \log \mathcal{L}(y_i^{\tau}; \nu_i, \sigma_i^2) &+ \sup_{\hat{\nu}, \hat{\sigma}^2} \sum_{\tau = s + 1}^t \log \mathcal{L}(y_i^{\tau}; \hat{\nu}, \hat{\sigma}^2) \nonumber \\ 
&- \sum_{\tau = 1}^t \log \mathcal{L}(y_i^{\tau}; \nu_i, \sigma_i^2),
\label{eq:psi}
\end{align}
where $\mathcal{L}(\cdot; \nu, \sigma^2)$ denotes the likelihood function of a Gaussian distribution with mean $\nu$ and variance $\sigma^2$. Note that $\Psi_i(s)$ essentially measures the gain in likelihood by explaining the observed data during $(s, t]$ with an alternative distribution. Based on the GLR statistic, the estimated change point is given as
\begin{equation}
\hat{t}_c(i) = \min \Bigl\{t \in [1, T]: \max_{s \in [1, t)} \Psi_i(s) \geq \eta \Bigr\},
\end{equation}
where $\eta$ is a positive constant. As a threshold, the value of $\eta$ determines the false alarm and miss rates of the GLR test. Herein, we derive the functional dependence between the threshold and the false alarm rate, so that $\eta$ can be chosen to achieve a desired false alarm probability,
\begin{equation}
\P\Bigl(\max_{s \in [1, t)} \Psi_i(s) \geq \eta \Big| \mathcal{H}_0 \Bigr) = \zeta.
\end{equation}

With the maximum likelihood estimates for Gaussian parameters,
\begin{equation} \label{eq:mle_gauss}
\hat{\nu}_i \doteq \frac{1}{t-s} \sum_{\tau = s + 1}^t y_i^{\tau}, \quad \hat{\sigma}_i^2 \doteq \frac{1}{t-s} \sum_{\tau = s + 1}^t (y_i^{\tau} - \hat{\nu}_i)^2,
\end{equation}
the test statistic under $\mathcal{H}_0$ can be formulated as
\begin{align}
&\Psi_i(s) = \sum_{\tau = s + 1}^t \log \mathcal{L}(y_i^{\tau}; \hat{\nu}_i, \hat{\sigma}_i^2) - \log \mathcal{L}(y_i^{\tau}; \nu_i, \sigma_i^2) \label{eq:psi_s_1} \\
&\quad = \sum_{\tau = s + 1}^t \log\left[\frac{\sigma_i}{\hat{\sigma}_i} \exp\Bigl(\frac{(y_i^{\tau} - \nu_i)^2}{2 \sigma_i^2} - \frac{(y_i^{\tau} - \hat{\nu}_i)^2}{2 \hat{\sigma}_i^2}\Bigr) \right] \\
&\quad = \frac{t-s}{2} \bigl(\log\sigma_i^2 - \log\hat{\sigma}_i^2 - 1\bigr) + \sum_{\tau = s + 1}^t \frac{(y_i^{\tau} - \nu_i)^2}{2 \sigma_i^2},
\label{eq:psi_s_2}
\end{align}
where we substitute $\hat{\sigma}_i^2$ in \eqref{eq:mle_gauss} to arrive at \eqref{eq:psi_s_2}. Applying a Taylor expansion of $\log\hat{\sigma}_i^2$ around $\sigma_i^2$, the test statistic is approximated as
\begin{align}
&\hspace{-6pt} 2\Psi_i(s) \approx (t-s) \biggl[\frac{1}{2} \Bigl(\frac{\hat{\sigma}_i^2 - \sigma_i^2}{\sigma_i^2}\Bigr)^2 - \frac{\hat{\sigma}_i^2}{\sigma_i^2}\biggr] + \sum_{\tau = s + 1}^t \frac{(y_i^{\tau} - \nu_i)^2}{\sigma_i^2} \\
&= \frac{t-s}{2} \Bigl(\frac{\hat{\sigma}_i^2}{\sigma_i^2} - 1\Bigr)^2 + \biggl[\frac{\sqrt{t-s}}{\sigma_i}(\hat{\nu_i} - \nu_i)\biggr]^2 \\
&= \biggl[\frac{(t-s) \hat{\sigma}_i^2 / \sigma_i^2 - (t-s)}{\sqrt{2(t-s)}}\biggr]^2 + \biggl[\frac{\sqrt{t-s}}{\sigma_i}(\hat{\nu_i} - \nu_i)\biggr]^2 \\
&\approx \biggl[\frac{(t-s) \hat{\sigma}_i^2 / \sigma_i^2 - (t-s-1)}{\sqrt{2(t-s-1)}}\biggr]^2 + \biggl[\frac{\sqrt{t-s}}{\sigma_i}(\hat{\nu_i} - \nu_i)\biggr]^2. \label{eq:approx_psi_2}
\end{align}
The sample mean and variance satisfy
\begin{align}
(t-s) \hat{\sigma}_i^2 / \sigma_i^2 \sim \chi^2(t-s-1), \quad \hat{\nu_i} \sim \Gsn\Bigl(\nu_i, \frac{\sigma_i^2}{t-s}\Bigr),
\end{align}
where $\chi^2(d)$ refers to the chi-squared distribution with $d$ degrees of freedom. If we approximate $\chi^2(t-s-1)$ by a Gaussian distribution, \eqref{eq:approx_psi_2} can be considered as the sum of the squares of two Gaussian random variables. Furthermore, these two variables are independent because the sample mean and sample variance of Gaussian distributions are independent \cite{rao1973linear}. Thus, we can approximate $2\Psi_i(s)$ by a chi-squared random variable with two degrees of freedom.

Note that $\Psi_i(s)$ is computed with samples in $(s, t]$, as shown in \eqref{eq:psi_s_1}. As a result, a strong positive correlation exists between the statistics calculated using overlapping subsets of the samples. For example, $\Psi_i(s_1)$ and $\Psi_i(s_2)$ are computed with overlapped samples in the interval $(\max(s_1,s_2), t]$. Therefore, we approximate the false alarm probability as
\begin{multline}
\P\Bigl(\max_{s \in [1, t)} \Psi_i(s) \geq \eta \Big| \mathcal{H}_0 \Bigr) \approx \\
\P\Bigl(2\Psi_i(s = 1) \geq 2\eta \Big| \mathcal{H}_0 \Bigr) \approx 1 - F_2(2\eta),
\end{multline}
where $F_2(\cdot)$ stands for the cumulative distribution function of $\chi^2(2)$. To achieve a false alarm rate of $\zeta$, we can set the threshold to be
\begin{equation}
\eta(\zeta) = F_2^{-1}(1-\zeta)/2.
\end{equation}
While our proposed method may not be optimal, we emphasize that the approach can detect changes at the sub-graph level, resulting in improved sample and computational efficiency.

Once a change is detected for a sub-graph at time $\hat{t}_c$, the edge weights associated with that sub-graph should be estimated only using observed values during $(\hat{t}_c, t]$. The proposed CSL-UCB algorithm combined with change detection is denoted as the \emph{CSL-UCB-CD} scheme.

\section{Numerical Results} \label{sec:NR}
In this section, we numerically evaluate the performance of the proposed schemes in terms of both graph identification and intervention selection. Both stationary and non-stationary causal structures are considered for evaluation. For each Monte Carlo run, the causal structure is randomly generated, with the edge weights randomly sampled from the uniform distribution $\mathcal{U}(-2, 2)$. The exogenous variables are independently sampled in each time step from the Gaussian distribution $\Gsn(1,1)$. To estimate the empirical mutual information, the $k$-nearest neighbor distances based approach is employed \cite{kraskov2004estimating}. For each set of parameters, we repeat the Monte Carlo run $M = 100$ times.\footnote{The code is available at \url{https://github.com/CalixPeng/Causal_Bandit}. }

\subsection{Performance for Causal Graph Identification} \label{subsec:PCGI}
We start by evaluating the causal graph learning ability of the proposed CSL scheme. For comparison, we consider the GOLEM algorithm from \cite{ng2020role}, the DAGMA algorithm from \cite{bello2022dagma} and the CSL-nw scheme, which refers to the proposed CSL scheme without edge weight regularization. The GOLEM algorithm provides an estimate of the observational weight matrix in step $t$ as 
\begin{equation}
\argmin_{\widehat{\bm{B}} \in \reals^{N \times N}} \Bigl\{-\log\mathcal{L}(\bm{X}^{1:t}; \widehat{\bm{B}}) + \beta_1 \bigl\lVert \widehat{\bm{B}} \bigr\rVert_1 + \beta_2 \bigl[\mathrm{tr}\bigl(e^{\widehat{\bm{B}} \circ \widehat{\bm{B}}}\bigr) - N \bigr]\Bigr\},
\end{equation}
where $\beta_1$ and $\beta_2$ are the penalty coefficients, $\circ$ denotes the Hadamard product and the negative log-likelihood is
\begin{align}
&-\log\mathcal{L}(\bm{X}^{1:t};\widehat{\bm{B}}) = \frac{N}{2} \log\biggl(\sum_{i=1}^N \sum_{\tau = 1}^t \bigl(x_i^{\tau} - \trans{\widehat{\bm{B}}}_i \bm{x}^{\tau}\bigr)^2 \biggr) \nonumber \\
&\hspace{140pt} - \log \bigl|\mathrm{det}(\bm{I} - \widehat{\bm{B}})\bigr|.
\end{align}
With a different characterization of acyclicity, the DAGMA algorithm estimates the weight matrix as
\begin{align}
&\argmin_{\widehat{\bm{B}} \in \reals^{N \times N}} \Bigl\{-\log\mathcal{L}(\bm{X}^{1:t}; \widehat{\bm{B}}) + \beta_1 \bigl\lVert \widehat{\bm{B}} \bigr\rVert_1 \nonumber \\ 
&\hspace{60pt} - \log\mathrm{det}(\beta_3 \bm{I} - \widehat{\bm{B}} \circ \widehat{\bm{B}}) + N \log \beta_3 \Bigr\},
\end{align}
where $\beta_3$ is the log-determinant coefficient.

\begin{figure}[!t]
\centering
\includegraphics[width=0.9\linewidth]{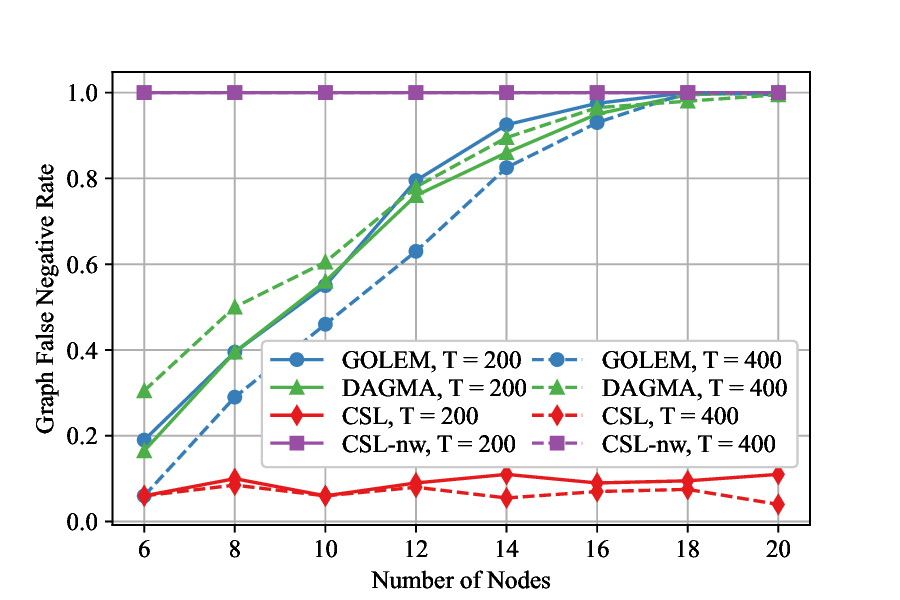}
\caption{Graph false negative rate as a function of graph sizes.}
\label{fig:graph_false_neg}
\end{figure}

Fig. \ref{fig:graph_false_neg} plots the graph FN rate as a function of graph sizes, with two sample sizes, $T = 200$ and $400$. The graph FN rate is defined as
\begin{equation}
\frac{1}{M} \sum_{m=1}^M \mathbb{I}\bigl(\exists (i, j) \in \mathcal{V} \times \mathcal{V}: \bm{B}_{ij}(m) \neq 0, \widehat{\bm{B}}_{ij}(m) = 0\bigr),
\end{equation}
where $\bm{B}(m)$ and $\widehat{\bm{B}}(m)$ represent the true and estimated edge weight matrices in the $m$-th Monte Carlo run. Observe that the CSL-nw scheme has a graph FN rate of $100\%$ in all cases, which indicates that without edge weight regularization, the estimated graphs always have FN errors. The GOLEM scheme performs slightly better than the DAGMA scheme, while both of them perform much worse than the proposed CSL scheme. This is due to the fact that GOLEM and DAGMA do not distinguish between FP and FN errors, while the CSL scheme is designed to explicitly consider the FN rate. On average, the CSL scheme achieves a graph FN rate that is lower than the GOLEM scheme by $81.1\%$ and the DAGMA scheme by $86.9\%$. Interestingly, given a fixed number of samples, as the graph size increases, the graph FN rates for both GOLEM and DAGMA increase rapidly. In contrast, The CSL scheme does not suffer much from the increased graph size, as learning sub-graphs has higher sample efficiency.

\begin{figure}[!t]
\centering
\includegraphics[width=0.9\linewidth]{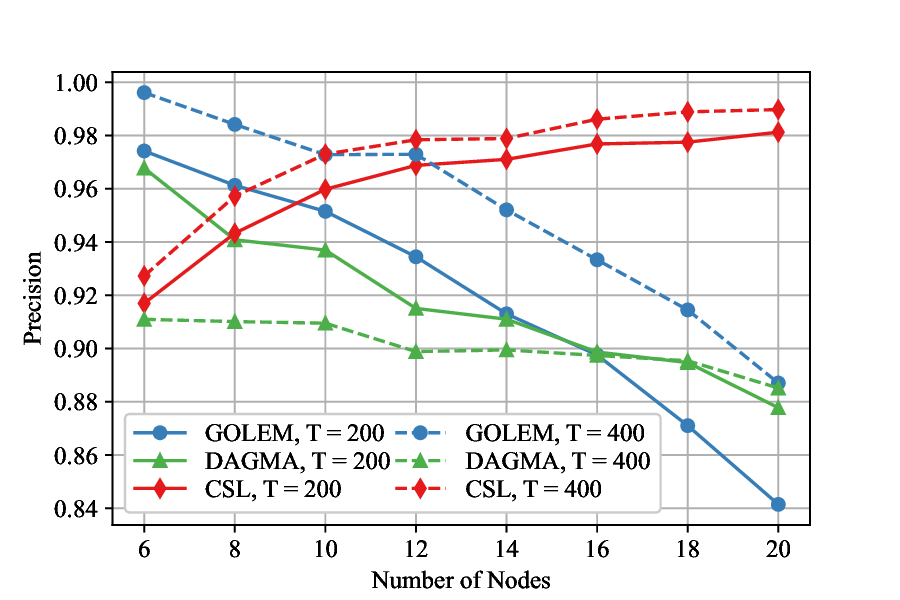}
\caption{Precision as a function of graph sizes.}
\label{fig:precision}
\end{figure}

\begin{figure}[!t]
\centering
\includegraphics[width=0.9\linewidth]{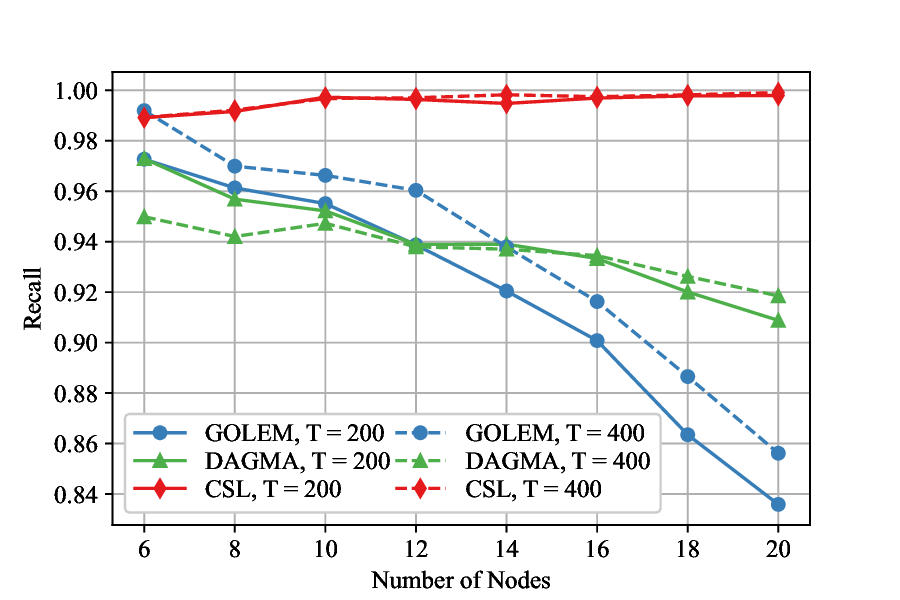}
\caption{Recall as a function of graph sizes.}
\label{fig:recall}
\end{figure}

To better understand the balance between false positives and false negatives, we consider precision and recall for evaluation, which are defined as 
\begin{align}
\mathrm{precision} &\doteq \frac{\sum_{i, j} \mathbb{I}\bigl(\bm{B}_{ij} \neq 0, \widehat{\bm{B}}_{ij} \neq 0 \bigr)}{\sum_{i,j} \mathbb{I}\bigl(\widehat{\bm{B}}_{ij} \neq 0 \bigr)}, \\
\mathrm{recall} &\doteq \frac{\sum_{i, j} \mathbb{I}\bigl(\bm{B}_{ij} \neq 0, \widehat{\bm{B}}_{ij} \neq 0 \bigr)}{\sum_{i,j} \mathbb{I}\bigl(\bm{B}_{ij} \neq 0 \bigr)}.
\end{align}
Note that precision and recall are inversely correlated with the numbers of FP and FN, respectively. Figure \ref{fig:precision} and \ref{fig:recall} plot precision and recall as functions of graph sizes. We observe that the proposed CSL scheme achieves a close to one recall by focusing on the reduction of FN errors. On average, the recall rates of GOLEM, DAGMA and CSL are $92.7\%$, $93.8\%$ and $99.6\%$. On the other hand, the CSL scheme does not perform significantly better than GOLEM and DAGMA in terms of precision, especially for small graphs. The average precision rates achieved by GOLEM, DAGMA and CSL are $93.5\%$, $90.9\%$ and $96.7\%$. We also observe that both precision and recall of the CSL scheme increase as the graph becomes larger, while GOLEM and DAGMA have the opposite trend. It suggests that for the CSL scheme, as $N$ increases, the number of mistaken edges increases more slowly than the totally number of edges, which increases quadratically in $N$.

\begin{figure}[!t]
\centering
\includegraphics[width=0.9\linewidth]{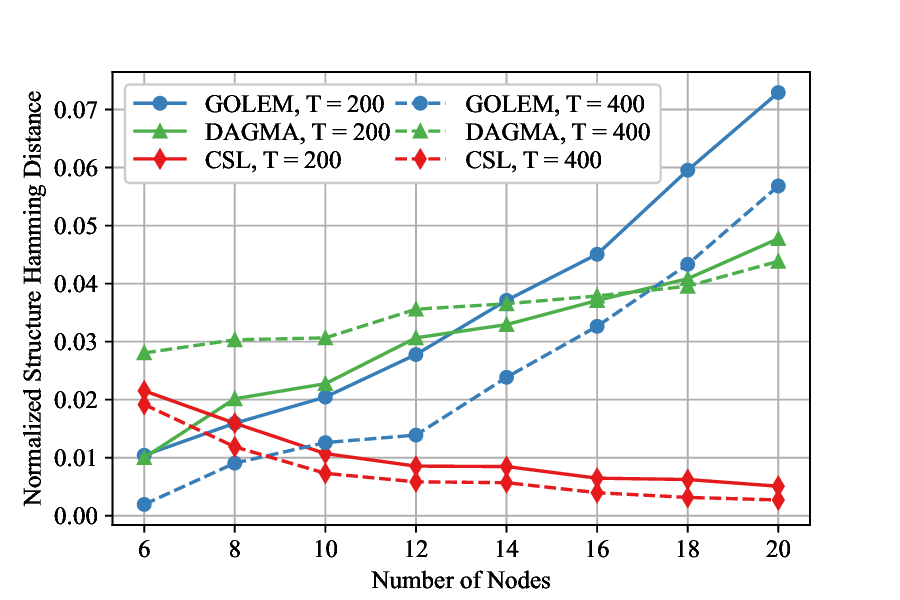}
\caption{Normalized Hamming distance as a function of graph sizes.}
\label{fig:hamming_dis}
\end{figure}

For completeness, we also include performance for both types of graph errors, measured by the normalized structural Hamming distance (NSHD), which is defined as \begin{align}
\mathrm{NSHD} \doteq \sum_{i, j} \Bigl[&\mathbb{I}\bigl(\bm{B}_{ij} = 0,  \widehat{\bm{B}}_{ij} \neq 0 \bigr) + \mathbb{I}\bigl(\bm{B}_{ij} \neq 0,  \widehat{\bm{B}}_{ij} = 0 \bigr) \nonumber \\
& - \mathbb{I}\bigl(\bm{B}_{ij} \neq 0,  \widehat{\bm{B}}_{ij} = 0, \widehat{\bm{B}}_{ji} \neq 0 \bigr) \Bigr]/ N^2.
\end{align}
In Fig. \ref{fig:hamming_dis}, the NSHD is plotted as a function of graph sizes. We observe that in terms of NSHD, the CSL scheme does not perform significantly better than the GOLEM and DAGMA schemes. Particularly when the graph is small ($N = 6$), DAGMA achieves the lowest normalized SHD for $200$ samples, while GOLEM achieves the lowest SHD for $400$ samples. Since both GOLEM and DAGMA learn the whole graph with gradient-based optimization, a sufficient amount of samples are necessary to achieve desirable performance. As the graph size increases, CSL offers strongly superior performance due to its need for fewer samples than the other methods.

\subsection{Performance for Stationary Causal Bandits} \label{subsec:PSCB}
\begin{figure}[!t]
\centering
\includegraphics[width=0.9\linewidth]{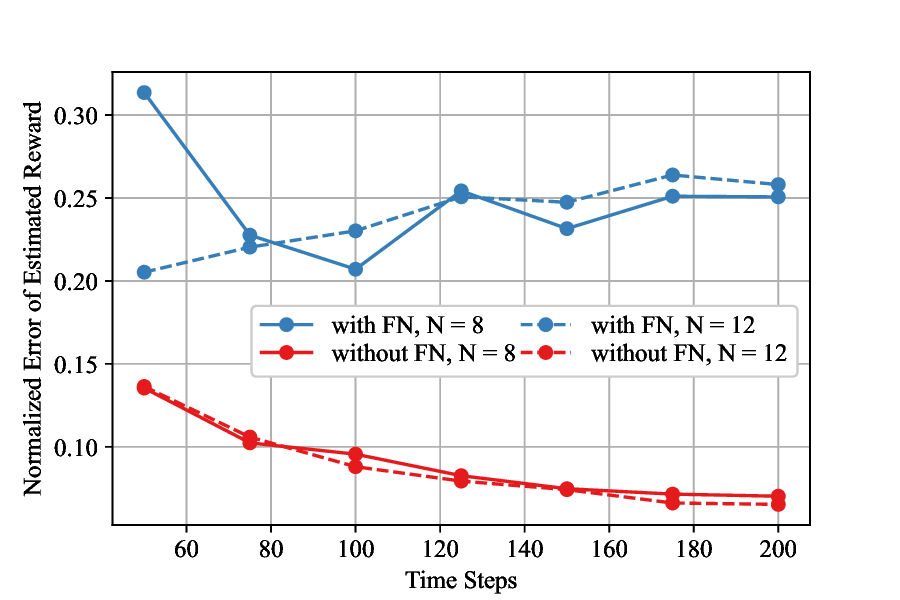}
\caption{Normalized error of estimated reward as a function of time steps.}
\label{fig:error_FN}
\end{figure}

Next, we consider the proposed CSL-UCB scheme with optimized intervention design in causal bandits. Since the goal is to maximize the cumulative reward, we start by revealing the impact of FN errors on reward estimation. To measure the impact, the estimation errors are classified into two groups based on whether FN error exists in the estimated graph. In Fig. \ref{fig:error_FN}, the normalized empirical reward error is plotted as a function of time for two cases: with or without FN errors. Without FN errors, the reward estimation error decreases monotonically as more samples are collected, which matches the analysis in Section \ref{subsec:AGE}. With the presence of FN errors, the estimation error does not decrease as more samples are collected. On average, the estimation error is $62.8\%$ lower in the absence of FN errors. The result is consistent with our claim that FN errors have a significant impact on reward maximization in causal bandits. The result also validates our choice to minimize the FN rate in causal graph identification.

To evaluate the CSL-UCB scheme, we focus on graphs with size $N = 10$ and simulate causal bandits over a horizon of $T = 1500$ time steps. To alleviate computation cost, the graphs are updated with new collected samples every $20$ time slots. For comparison, we first consider the vanilla UCB algorithm \cite{sutton2018reinforcement}, which selects an intervention in each step as
\begin{equation}
\argmax_{\bm{a}} \left\{\frac{\sum_{\tau = 1}^t \mathbb{I}(\bm{a}^{\tau}=\bm{a}) X_N^{\tau}}{\sum_{\tau = 1}^t \mathbb{I}(\bm{a}^{\tau}=\bm{a})} + \alpha' \sqrt{\frac{\ln t}{\sum_{\tau = 1}^t \mathbb{I}(\bm{a}^{\tau}=\bm{a})}} \right\},
\end{equation}
where $\alpha'$ is the parameter that controls the exploration level. Notice that the vanilla UCB algorithm does not exploit the causal structure and its sample complexity scales exponentially as $2^N$ \cite{lattimore2020bandit}. Another scheme we consider for comparison is the GOLEM \cite{ng2020role} algorithm based decision-making for MABs, denoted as the GOLEM-MAB scheme. The horizon is divided into two parts, such that causal structure identification is conducted in the first part, while rewards are collected in the second part by exploiting the learned causal graph. Note that since GOLEM identifies the whole graph, intervention should be fixed as $\bm{a} = \bm{0}$ for learning $\bm{B}$ and $\bm{a} = \bm{1}$ for learning $\bm{B}'$. Lastly, we employ the LinSEM-TS scheme proposed in \cite{varici2023causal}, which provides an upper bound on performance, as it utilizes knowledge of the causal topology.

\begin{figure}[!t]
\centering
\includegraphics[width=0.9\linewidth]{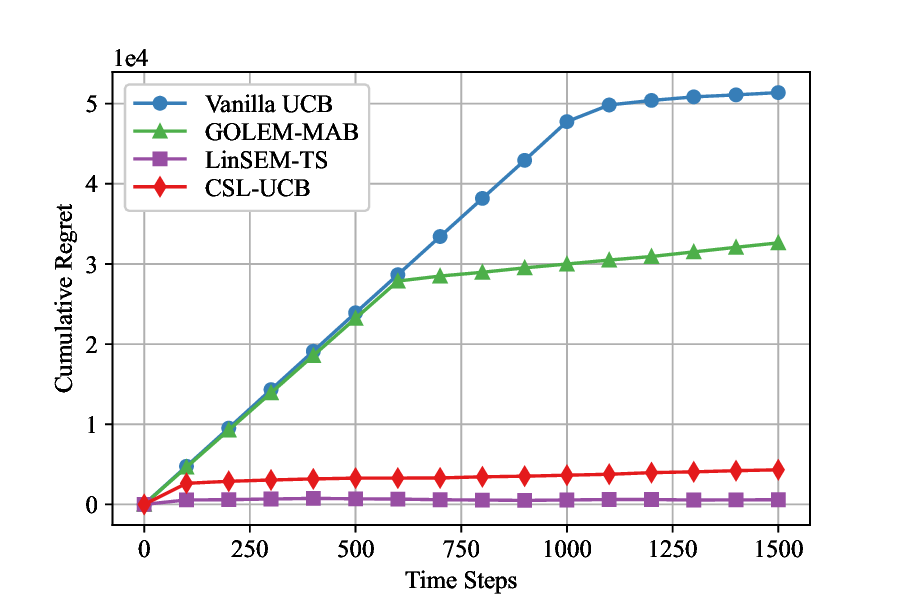}
\caption{Cumulative regret of stationary bandits as a function of time steps.}
\label{fig:cum_regret}
\end{figure}

\begin{figure}[!t]
\centering
\includegraphics[width=0.9\linewidth]{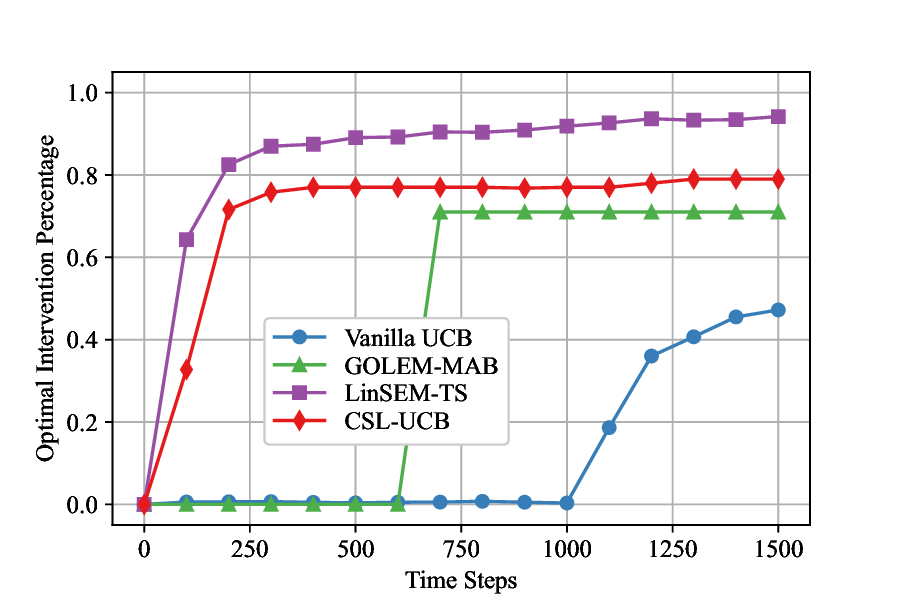}
\caption{Optimal percentage of stationary bandits as a function of time steps.}
\label{fig:opt_percent}
\end{figure}

Figure \ref{fig:cum_regret} plots the cumulative regret as a function of time for four different algorithms and the associated probability of selecting the optimal intervention is provided in Figure \ref{fig:opt_percent}. The vanilla UCB algorithm gains information about the optimal intervention after exploring every possible intervention at $t = 2^N = 1024$. By the end of the horizon, vanilla UCB selects the optimal intervention $47.2\%$ of the time. GOLEM-MAB tries to identify the causal graph in the first $600$ steps, and exploits this knowledge to achieve an optimal intervention selection ratio of $71.0\%$. The value of $t = 600$ was selected via trial and error to optimize the performance of GOLEM-MAB. The proposed CSL-UCB scheme gains and leverages causal knowledge in an alternating manner, yielding an optimal intervention selection ratio of $79.0\%$. With topology knowledge, LinSEM-TS is able to select the optimal intervention with probability $94.2\%$. In terms of cumulative regret, the CSL-UCB scheme achieves a $91.6\%$ improvement compared with vanilla UCB and an $86.8\%$ improvement compared with GOLEM-MAB. The LinSEM-TS algorithm can further reduce $86.6\%$ of the cumulative regret the CSL-UCB scheme achieves when the causal structure is fully known.

An interesting observation is that, after $600$ time steps for graph identification ($300$ for each, $\bm{B}$ and $\bm{B}'$), the graphs estimated by GOLEM are close to the ground truth with respect to the SHD. However, even rejecting one existing edge (one false negative error) could result in a significantly different expected reward. Since edges are highly possible to be part of the causal flows that influence the reward node, mistakenly rejecting an edge can cut off one or multiple relevant causal flows and thus significantly impact reward estimation. The relatively poor performance of GOLEM-MAB is not due to its generic graph identification ability, but the equal treatment of all errors and its low sample efficiency. We emphasize again that causal bandit optimization and causal graph identification are two different problems.

\subsection{Performance for Non-stationary Causal Bandits} \label{subsec:PNCB}
Lastly, we evaluate the reward-earning ability of the proposed CSL-UCB-CD scheme in non-stationary causal bandits. Structure change occurs twice at $t \in \{1000, 2000\}$, where a random subset of the causal mechanisms are regenerated. For comparison, the vanilla UCB algorithm can adjust its estimate of the rewards after structure changes using newly collected samples. The GOLEM-MAB algorithm combined with a change detection mechanism is denoted as GOLEM-MAB-CD, which must re-learns the whole causal graph when a change is detected. The LinSEM-TS algorithm initially has full knowledge of the causal structure, but is unaware of the upcoming changes. 

\begin{figure}[!t]
\centering
\includegraphics[width=0.9\linewidth]{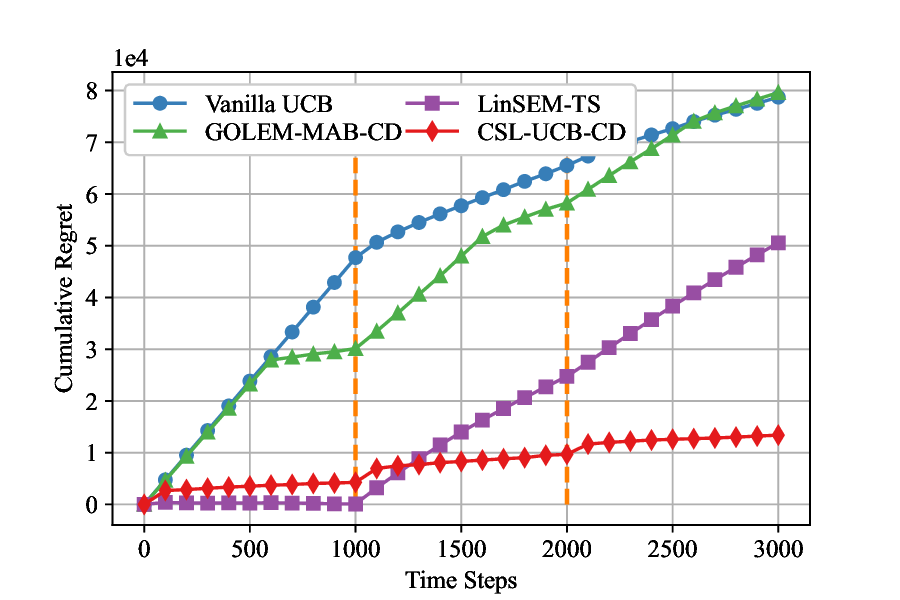}
\caption{Cumulative regret of non-stationary bandits as a function of time steps. Structure change occurs twice, as marked by the dashed lines. }
\label{fig:cum_regret_c}
\end{figure}

\begin{figure}[!t]
\centering
\includegraphics[width=0.9\linewidth]{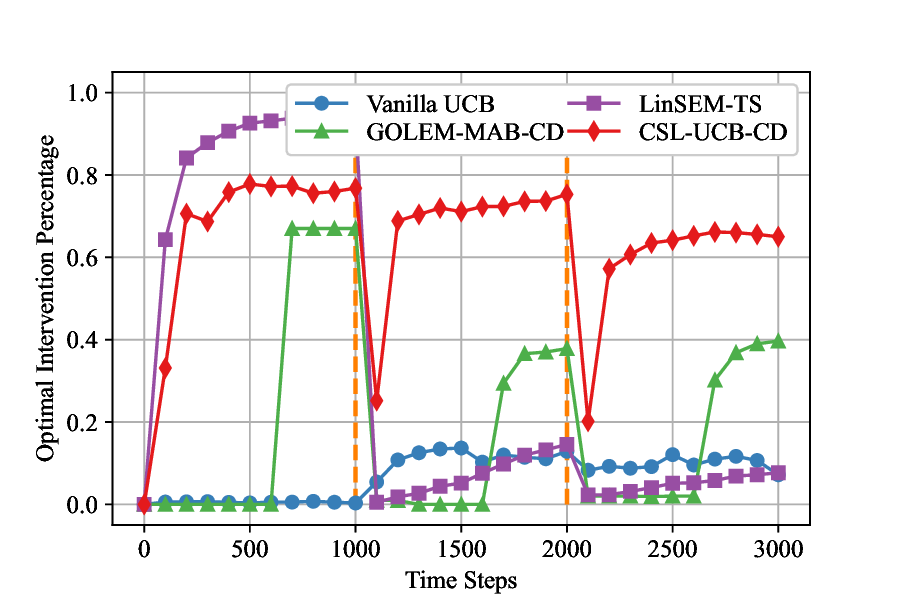}
\caption{Optimal percentage of non-stationary bandits as a function of time steps. Structure change occurs twice, as marked by the dashed lines. }
\label{fig:opt_percent_c}
\end{figure}

In Fig. \ref{fig:cum_regret_c}, the cumulative regret is plotted as a function of time. The percentage of selecting the optimal intervention as a function of time is provided in Fig. \ref{fig:opt_percent_c}. On average, the change detection mechanism is able to detect $89.3\%$ of the sub-graph changes, with an average delay of $13.1$ slots. The cumulative regret achieved by the CSL-UCB-CD scheme is $83.0\%$ lower than vanilla UCB, $83.2\%$ lower than GOLEM-MAB-CD and $73.5\%$ lower than LinSEM-TS. Although LinSEM-TS obtains the lowest regret before structure change, its performance degrades dramatically when the causal knowledge becomes incorrect.

The vanilla UCB scheme performs the worst, not due to structure changes but rather, due to the lack of samples. Averaged over time, the optimal intervention ratio of the vanilla UCB scheme is only $7.0\%$. As illustrated in Fig. \ref{fig:opt_percent_c}, the optimal percentage of GOLEM-MAB-CD drops to $0\%$ after each change-point, which indicates that it is able to detect the changes and start to re-learn the causal graph. However, as learning the whole graph requires a large amount of samples, GOLEM-MAB-CD is only able to select the optimal intervention $18.3\%$ of the time. Since LinSEM-TS suffers from incorrect causal knowledge caused by structure changes, it selects the optimal intervention only $32.6\%$ of the time. The CSL-UCB-CD scheme combines sub-graph change detection with sub-graph learning, and select the optimal intervention $63.8\%$ of the time.

Compared with the stationary case, it is observed that the regret gap between vanilla UCB and GOLEM-MAB-CD in non-stationary bandits is smaller. The major reason is that, utilizing samples from the previous model, vanilla UCB implicitly assumes those samples contain information about the current structure. This is true, because the changes occur only for a subset of the causal mechanisms. We also observe that CSL-UCB-CD performs better than GOLEM-MAB-CD mainly due to its fast recovery from structure changes. This fast recovery property comes from the fact that re-learning sub-graphs requires fewer samples than re-learning whole graphs.

\section{Conclusions} \label{sec:CC}
In this paper, we investigate the causal bandit problem with no knowledge of the causal graph topology and the interventional distribution. Distinguishing between the two types of error, a sub-graph learning-based causal identification approach is proposed, which makes efficient use of the limited data. Due to learning sub-graphs versus entire graphs, the complexity of the proposed method is significantly lower than traditional methods. Analysis reveals that optimal intervention design is much more sensitive to false negative errors in graph identification. This feature is directly taken into consideration for the sub-graph learning algorithm. To balance exploration and exploitation, we analyze and propose an uncertainty bound. Further, sub-graph level change detection mechanism is proposed, which enables the proposed approach to adapt to non-stationary bandits. Numerical results show that the overall approach is able to identify the optimal intervention with much lower sample and computational complexity than previous methods. Furthermore, performance is improved over prior approaches due to the direct consideration of false negative errors. The proposed approach requires $67\%$ fewer samples to learn the causal structure and achieves a lower cumulative regret by $85\%$, compared to the existing schemes.

\appendices
\section{Proof of Proposition \ref{prop:I_decomp}} \label{proof:prop_I_decomp}
\begin{IEEEproof}
We start by proving a general statement for continuous random variables. That is, the following inequality holds,
\begin{equation}
I(Z_1 + Z_2; Z_3 + Z_4) \leq I(Z_1; Z_3) + I(Z_2;Z_4),
\end{equation}
if $Z_1 \perp Z_2, Z_4$ and $Z_3 \perp Z_2, Z_4$. To prove that, we first employ the data processing inequality \cite{cover1999elements} to bound the mutual information as
\begin{align}
I(Z_1 + Z_2; Z_3 + Z_4) &\leq I(Z_1, Z_2; Z_3 + Z_4) \\ 
&\leq I(Z_1, Z_2; Z_3, Z_4).
\end{align}
Then, based on the definition of mutual information, we have
\begin{align}
&\hspace{-12pt} I(Z_1, Z_2; Z_3, Z_4) \nonumber \\
&= \int f(z_1,z_2,z_3,z_4) \log \frac{f(z_1,z_2,z_3,z_4)}{f(z_1,z_2) f(z_3,z_4)} d\bm{z}
\end{align}
\begin{align}
&= \int f(z_1,z_3) f(z_2,z_4) \log \frac{f(z_1,z_3)}{f(z_1)f(z_3)} d\bm{z} \nonumber \\
&\hspace{40pt} + \int f(z_1,z_3) f(z_2,z_4) \log \frac{f(z_2,z_4)}{f(z_2)f(z_4)} d\bm{z} \\
&= \int f(z_1,z_3) \log \frac{f(z_1,z_3)}{f(z_1)f(z_3)} dz_1 dz_3 \cdot 1  \nonumber \\
&\hspace{40pt} + \int f(z_2,z_4) \log \frac{f(z_2,z_4)}{f(z_2)f(z_4)} dz_2 dz_4 \cdot 1 \\
&= I(Z_1; Z_3) + I(Z_2;Z_4).
\end{align}
Given this inequality and the fact that the exogenous variables are independent, the substitution 
\begin{align}
Z_1 &= \sum_{l \not\in \mathcal{A}_i(a_i)} \bigl[\widehat{\bm{C}}_{\bm{a}}^t\bigr]_{li} \epsilon_l - \tilde{\epsilon}_i, \nonumber \\
Z_2 &= \sum_{k \in \mathcal{A}_i(a_i)} \bigl(\bigl[\widehat{\bm{C}}_{\bm{a}}^t\bigr]_{ki} - \left[\bm{C}_{\bm{a}}\right]_{ki}\bigr) \epsilon_k, \nonumber \\
Z_3 &= \sum_{l \not\in \mathcal{A}_i(a_i)} \left[\bm{C}_{\bm{a}}\right]_{lj} \epsilon_l, \quad Z_4 = \sum_{k \in \mathcal{A}_i(a_i)} \left[\bm{C}_{\bm{a}}\right]_{kj} \epsilon_k
\end{align}
completes the proof.
\end{IEEEproof}

\section{Proof of Proposition \ref{prop:graph_error}} \label{proof:prop_graph_error}
\begin{IEEEproof}
Consider two sets $\mathcal{Q}$, $\mathcal{S}$ such that $\mathcal{Q} \subset \mathcal{S}$ and the observed values of nodes up to step $t$ satisfy
\begin{align}
\bm{X}^{1:t}_i(a_i) &= \trans{\left[\bm{B}_{\bm{a}}\right]}_{i,\mathcal{Q}} \bm{X}^{1:t}_{\mathcal{Q}}(a_i) + \trans{\bm{0}} \bm{X}^{1:t}_{\mathcal{S} \backslash \mathcal{Q}}(a_i) + \bm{\epsilon}^{1:t}_i(a_i) \\
&= \trans{\left[\bm{B}_{\bm{a}}\right]}_{i,\mathcal{S}} \bm{X}^{1:t}_{\mathcal{S}}(a_i) + \bm{\epsilon}^{1:t}_i(a_i)
\end{align}
where $\bm{0}$ is a vector of zeros and $\mathcal{S} \backslash \mathcal{Q}$ denotes the set difference of $\mathcal{S}$ and $\mathcal{Q}$. The expected vector of weights associated with nodes in $\mathcal{S}$, is given by the MMSE estimation as
\begin{align}
&\hspace{-4pt} \E\bigl[\bigl[\widehat{\bm{B}}_{\bm{a}}^t\bigr]_{i,\mathcal{S}} \big| \bm{X}^{1:t}_{\mathcal{S}}(a_i)\bigr] = \E\bigl[\bigl[\bm{X}^{1:t}_{\mathcal{S}}(a_i) \trans{\bm{X}^{1:t}_{\mathcal{S}}(a_i)}\bigr]^{-1} \bm{X}^{1:t}_{\mathcal{S}}(a_i) \nonumber \\
&\hspace{38pt} \cdot (\trans{\bm{X}^{1:t}_{\mathcal{S}}(a_i)} \left[\bm{B}_{\bm{a}}\right]_{i,\mathcal{S}} + \trans{\tilde{\bm{\epsilon}}^{1:t}_i(a_i)}) \big| \bm{X}^{1:t}_{\mathcal{S}}(a_i)\bigr] \\
&= \left[\bm{B}_{\bm{a}}\right]_{i,\mathcal{S}} + \bm{0} \quad = \left[\bm{B}_{\bm{a}}\right]_{i,\mathcal{S}}. \label{eq:Ba_fp}
\end{align}
With the substitution $\mathcal{Q} = \mathcal{P}_i(a_i)$, $\mathcal{S} = \widehat{\mathcal{P}}_i(a_i)$, we obtain the effect of including non-existent parents (false positive error). Equation \eqref{eq:Ba_fp} holds because the exogenous variable $\tilde{\epsilon}_i$ is independent of the variables associated with the nodes in $\mathcal{S} = \widehat{\mathcal{P}}_i(a_i)$. Without a false negative error, $\widehat{\mathcal{P}}_i(a_i)$ should not contain node $i$ and its descendants.

Next, consider the case that the values are generated as
\begin{equation}
\bm{X}^{1:t}_i(a_i) = \trans{\left[\bm{B}_{\bm{a}}\right]}_{i,\mathcal{Q}} \bm{X}^{1:t}_{\mathcal{Q}}(a_i) + \trans{\left[\bm{B}_{\bm{a}}\right]}_{i,\mathcal{S} \backslash \mathcal{Q}} \bm{X}^{1:t}_{\mathcal{S} \backslash \mathcal{Q}}(a_i) + \bm{\epsilon}^{1:t}_i(a_i).
\end{equation}
The vector of expected weights associated with $\mathcal{Q}$ is given by the MMSE estimation as
\begin{align}
&\hspace{-4pt} \E\bigl[\bigl[\widehat{\bm{B}}_{\bm{a}}^t\bigr]_{i,\mathcal{Q}} \big| \bm{X}^{1:t}_{\mathcal{Q}}(a_i)\bigr] = \E \bigl[\bigl[\bm{X}^{1:t}_{\mathcal{Q}}(a_i) \trans{\bm{X}^{1:t}_{\mathcal{Q}}(a_i)}\bigr]^{-1} \bm{X}^{1:t}_{\mathcal{Q}}(a_i) \nonumber \\ 
&\hspace{24pt} \bigl(\trans{\bm{X}^{1:t}_{\mathcal{Q}}(a_i)} \left[\bm{B}_{\bm{a}}\right]_{i,\mathcal{Q}} + \trans{\bm{X}^{1:t}_{\mathcal{S} \backslash \mathcal{Q}}(a_i)} \left[\bm{B}_{\bm{a}}\right]_{i,\mathcal{S} \backslash \mathcal{Q}} + \nonumber \\
&\hspace{138pt} \trans{\tilde{\bm{\epsilon}}^{1:t}_i(a_i)} \bigr) \big| \bm{X}^{1:t}_{\mathcal{Q}}(a_i)\bigr] \\
&\quad = \left[\bm{B}_{\bm{a}}\right]_{i,\mathcal{Q}} + \bigl[\bm{X}^{1:t}_{\mathcal{Q}}(a_i) \trans{\bm{X}^{1:t}_{\mathcal{Q}}(a_i)}\bigr]^{-1} \E\bigl[\bm{X}^{1:t}_{\mathcal{Q}}(a_i) \cdot \nonumber \\ 
&\hspace{84pt} \trans{\bm{X}^{1:t}_{\mathcal{S} \backslash \mathcal{Q}}(a_i)} \left[\bm{B}_{\bm{a}}\right]_{i,\mathcal{S} \backslash \mathcal{Q}}\big| \bm{X}^{1:t}_{\mathcal{Q}}(a_i)\bigr].
\end{align}
Note that having FN errors is equivalent to have $\widehat{\mathcal{P}}_i(a_i) \subset \mathcal{P}_i(a_i)$. Thus with the substitution $\mathcal{Q} = \widehat{\mathcal{P}}_i(a_i)$, $\mathcal{S} = \mathcal{P}_i(a_i)$, we obtain the effect of FN errors (rejecting actual edges).
\end{IEEEproof}

\section{Proof of Lemma \ref{lem:moment_bound}} \label{proof:lem_moment_bound}
\begin{IEEEproof}
When there is no false negative error, each weight error vector follows a multivariate normal distribution according to the theory of linear least-square estimation, 
\begin{equation}
\left[\Delta\bm{B}_{\bm{a}}^t\right]_i \sim \mathcal{N}(\bm{0}, \bm{\Phi}_{\bm{a}}^t(i)),
\end{equation}
where the covariance matrix can be decomposed as $\bm{\Phi}_{\bm{a}}^t(i) = \bm{Q}_i \bm{\Lambda}_i \trans{\bm{Q}}_i$. With the orthonormal matrix $\bm{Q}_i$, we have 
\begin{equation}
\left\lVert \left[\Delta\bm{B}_{\bm{a}}^t\right]_i \right\rVert_2 = \left\lVert \trans{\bm{Q}}_i \left[\Delta\bm{B}_{\bm{a}}^t\right]_i \right\rVert_2, \ \trans{\bm{Q}}_i \left[\Delta\bm{B}_{\bm{a}}^t\right]_i \sim \mathcal{N}(\bm{0}, \bm{\Lambda}_i).
\label{eq:weight_error_vec}
\end{equation}
Since the squared norm of $\trans{\bm{Q}}_i \left[\Delta\bm{B}_{\bm{a}}^t\right]_i$ is a weighted sum of independent squared Gaussian variables, it can be bounded from above as
\begin{align}
&\left\lVert \trans{\bm{Q}}_i \left[\Delta\bm{B}_{\bm{a}}^t\right]_i \right\rVert_2^2 \leq \lambda_{\max}(\bm{\Lambda}_i) \bigl\lVert \bm{\Lambda}_i^{-1/2} \trans{\bm{Q}}_i \left[\Delta\bm{B}_{\bm{a}}^t\right]_i \bigr\rVert_2^2 \\
&\hspace{60pt} = \lambda_{\max}(\bm{\Phi}_{\bm{a}}^t(i)) \bigl\lVert \bm{\Lambda}_i^{-1/2} \trans{\bm{Q}}_i \left[\Delta\bm{B}_{\bm{a}}^t\right]_i \bigr\rVert_2^2.
\label{eq:chi_bound}
\end{align}
Notice that $\bigl\lVert \bm{\Lambda}_i^{-1/2} \trans{\bm{Q}}_i \left[\Delta\bm{B}_{\bm{a}}^t\right]_i \bigr\rVert_2^2$ follows a Chi-squared distribution with $N$ degrees of freedom, with the $m$-th moment defined as
\begin{equation}
\E\left[\left(\bigl\lVert \bm{\Lambda}_i^{-1/2} \trans{\bm{Q}}_i \left[\Delta\bm{B}_{\bm{a}}^t\right]_i \bigr\rVert_2^2 \right)^m\right] = 2^m \frac{\Gamma(m+N/2)}{\Gamma(N/2)},
\label{eq:chi_moment}
\end{equation}
where $\Gamma(\cdot)$ represents the Gamma function. Thus we can upper bound the $m$-th moment of the error norm as
\begin{align}
&\E \Bigl[\bigl\lVert \left[\Delta\bm{B}_{\bm{a}}^t\right]_i \bigr\rVert_2^m \Bigr] \stackrel{(a)}{\leq} \sqrt{\E \Bigl[\left(\bigl\lVert \left[\Delta\bm{B}_{\bm{a}}^t\right]_i \bigr\rVert_2^2\right)^m \Bigr]} \\
&\hspace{44pt} \stackrel{(b)}{\leq} \sqrt{\lambda_{\max}\left(\bm{\Phi}_{\bm{a}}^t(i)\right)^m 2^m \frac{\Gamma(m+N/2)}{\Gamma(N/2)}} \\
&\hspace{44pt} \stackrel{(c)}{\leq} m! \sqrt{\frac{4N}{N+2}} \left[\lambda_{\max}(\bm{\Phi}_{\bm{a}}^t(i)) \frac{N+2}{4} \right]^{m/2},
\end{align}
where $(a)$ comes from Jensen's inequality while $(b)$ is a direct consequence of \eqref{eq:weight_error_vec}, \eqref{eq:chi_bound} and \eqref{eq:chi_moment}. To show $(c)$, we first rewrite the quotient of the Gamma function as
\begin{align}
2^m \frac{\Gamma(m+N/2)}{\Gamma(N/2)} &= N (N+2) \cdots (N+2m-2) \\
&= (m!)^2 \cdot \prod_{k=1}^m \frac{N+2k-2}{k^2}.
\end{align}
When $m \geq 2$, we have the following inequalities
\begin{equation}
\frac{N + 2k - 2}{k^2} = \frac{N-2}{k^2} + \frac{2}{k} \leq \frac{N+2}{4}, \quad \forall k \geq 2,
\end{equation}
\begin{equation}
2^m \frac{\Gamma(m+N/2)}{\Gamma(N/2)} \leq (m!)^2 \cdot N \cdot \left(\frac{N+2}{4}\right)^{m-1},
\end{equation}
which shows $(c)$ and thus completes the proof.
\end{IEEEproof}

\section{Proof of Lemma \ref{lem:sv_bound}} \label{proof:lem_sv_bound}
\begin{IEEEproof}
For a specific intervention $\bm{a}$, consider the dilation \cite{paulsen2002completely} of the weight error matrix,
\begin{equation}
\bm{H}(\Delta\bm{B}_{\bm{a}}^t) \doteq \begin{bmatrix}
\bm{0} & \Delta\bm{B}_{\bm{a}}^t \\
\trans{\left(\Delta\bm{B}_{\bm{a}}^t\right)} & \bm{0}
\end{bmatrix}.
\end{equation}
An important property is that the dilation preserves spectral information:
\begin{equation}
\lambda_{\max}(\bm{H}(\Delta\bm{B}_{\bm{a}}^t)) = \sigma_{\max}(\bm{H}(\Delta\bm{B}_{\bm{a}}^t)) = \sigma_{\max}(\Delta\bm{B}_{\bm{a}}^t),
\end{equation}
Since the weight matrix consists of independently learned columns, we decompose the dilation as
\begin{equation}
\bm{H}(\Delta\bm{B}_{\bm{a}}^t) = \sum_{i=1}^N \bm{H}_i, \quad \bm{H}_i \doteq \begin{bmatrix}
\bm{0} & \left(\Delta\bm{B}_{\bm{a}}^t\right)_i \\
\trans{\left(\Delta\bm{B}_{\bm{a}}^t\right)}_i & \bm{0}
\end{bmatrix},
\end{equation}
where the $N \times N$ matrix $\left(\Delta\bm{B}_{\bm{a}}^t\right)_i$ is composed of the $i$-th column of $\Delta\bm{B}_{\bm{a}}^t$ and zero everywhere else. Applying block matrix multiplication repeatedly, we have
\begin{equation}
\bm{H}_i^2 = \begin{bmatrix}
\left(\Delta\bm{B}_{\bm{a}}^t\right)_i \trans{\left(\Delta\bm{B}_{\bm{a}}^t \right)}_i & \bm{0} \\
\bm{0} & \trans{(\Delta\bm{B}_{\bm{a}}^t)}_i \left(\Delta\bm{B}_{\bm{a}}^t\right)_i
\end{bmatrix},
\end{equation}
\begin{equation}
\bm{H}_i^m = \begin{cases}
\bigl\lVert \left[\Delta\bm{B}_{\bm{a}}^t\right]_i \bigr\rVert_2^{m-1} \cdot \bm{H}_i, & m \ \text{is odd} \\
\bigl\lVert \left[\Delta\bm{B}_{\bm{a}}^t\right]_i \bigr\rVert_2^{m-2} \cdot \bm{H}_i^2, & m \ \text{is even}
\end{cases}.
\end{equation}
We further observe that
\begin{equation}
\lambda_{\max}(\bm{H}_i) = \bigl\lVert \left[\Delta\bm{B}_{\bm{a}}^t\right]_i \bigr\rVert_2, \ \lambda_{\max}(\bm{H}_i^2) = \bigl\lVert \left[\Delta\bm{B}_{\bm{a}}^t\right]_i \bigr\rVert_2^2,
\end{equation}
which enables us to draw a general conclusion that
\begin{equation}
\bm{H}_i^m \preceq \bigl\lVert \left[\Delta\bm{B}_{\bm{a}}^t\right]_i \bigr\rVert_2^m \cdot \bm{I},
\end{equation}
where $\preceq$ represents the matrix semi-definite ordering. Further, taking expectation and applying Lemma \ref{lem:moment_bound}, we have
\begin{align}
\E[\bm{H}_i] &= \bm{0} \\
\E[\bm{H}_i^m] &\preceq m! \sqrt{\frac{4N}{N+2}} \left[\lambda_{\max}(\bm{\Phi}_{\bm{a}}^t(i)) \frac{N+2}{4} \right]^{\frac{m}{2}} \bm{I}, \ m \geq 2.
\end{align}
These properties enable us to apply the matrix Bernstein inequality for the sub-exponential case \cite{tropp2012user} and obtain
\begin{align}
&\P\{\sigma_{\max}(\Delta\bm{B}_{\bm{a}}^t) \geq \xi \} = \P\{\lambda_{\max}(\bm{H}(\Delta\bm{B}_{\bm{a}}^t)) \geq \xi \} \\
&= \P\Bigl\{\lambda_{\max}\Bigl(\sum_{i=1}^N \bm{H}_i\Bigr) \geq \xi \Bigr\} \\
&\leq 2N \exp\left[-\frac{\xi^2}{4\sqrt{N(N+2)} \cdot \sum_{i=1}^N \lambda_{\max}(\bm{\Phi}_{\bm{a}}^t(i))} \right],
\end{align}
which is valid when
\begin{equation}
\xi \leq \sqrt{\frac{4N}{\max_i \lambda_{\max}(\bm{\Phi}_{\bm{a}}^t(i))}} \cdot \sum_{i=1}^N \lambda_{\max}(\bm{\Phi}_{\bm{a}}^t(i)).
\end{equation}
Lastly, we denote the probability bound by $\delta$ and solve $\xi$ from
\begin{equation}
\delta = 2N \exp\left[-\frac{\xi^2}{4\sqrt{N(N+2)} \cdot \sum_{i=1}^N \lambda_{\max}(\bm{\Phi}_{\bm{a}}^t(i))} \right],
\end{equation}
to arrive at the desired inequality.
\end{IEEEproof}

\section{Proof of Theorem \ref{thm:conf_bound}} \label{proof:thm_conf_bound}
\begin{IEEEproof}
First, recognize that the error of estimated reward is the last element of $\Delta\bm{\mu}_{\bm{a}}^t$. To understand the behavior of $\Delta\bm{\mu}_{\bm{a}}^t$, consider the following two linear systems,
\begin{align}
\trans{(\bm{I}-\bm{B_a})} \bm{\mu}_{\bm{a}} &= \bm{\nu}, \label{eq:LS_1} \\ 
\trans{(\bm{I}-\widehat{\bm{B}}_{\bm{a}}^t)} \hat{\bm{\mu}}_{\bm{a}}^t &= \bm{\nu} \label{eq:LS_2},
\end{align}
where \eqref{eq:LS_1} and \eqref{eq:LS_2} represent the true and estimated systems, respectively. Based on the Woodbury matrix identity \cite{higham2002accuracy}, we have
\begin{equation}
(\bm{I}-\widehat{\bm{B}}_{\bm{a}}^t)^{-1} = (\bm{I}-\bm{B}_{\bm{a}})^{-1} + (\bm{I}-\bm{B}_{\bm{a}})^{-1} \Delta\bm{B_a}^t (\bm{I}-\widehat{\bm{B}}_{\bm{a}}^t)^{-1}.
\label{eq:TE}
\end{equation}
Combining \eqref{eq:LS_2} with \eqref{eq:TE}, we can express $\Delta\bm{\mu}_{\bm{a}}$ as
\begin{align}
\Delta\bm{\mu}_{\bm{a}}^t &= -\bm{\mu}_{\bm{a}} + \invtrans{(\bm{I}-\bm{B}_{\bm{a}})} \bm{\nu} + \nonumber \\
& \hspace{45pt} \invtrans{(\bm{I}-\widehat{\bm{B}}_{\bm{a}}^t)} \trans{(\Delta\bm{B_a}^t)} \invtrans{(\bm{I}-\bm{B_a})} \bm{\nu} \\
&= \invtrans{(\bm{I}-\widehat{\bm{B}}_{\bm{a}}^t)} \trans{(\Delta\bm{B_a}^t)} \bm{\mu}_{\bm{a}}, \label{eq:del_mu}
\end{align}
where \eqref{eq:del_mu} is a direct consequence of \eqref{eq:LS_1}. Now, we can bound the error of estimated reward as
\begin{align}
&\left|\left[\Delta\bm{\mu}_{\bm{a}}^t\right]_N\right| = \left|\trans{\bigl[(\bm{I}-\widehat{\bm{B}}_{\bm{a}}^t)^{-1}\bigr]}_N \trans{(\Delta\bm{B_a}^t)} \bm{\mu}_{\bm{a}}\right| \\
&\hspace{24pt} \leq \left\lVert \bigl[(\bm{I}-\widehat{\bm{B}}_{\bm{a}}^t)^{-1}\bigr]_N \right\rVert_2 \cdot \sigma_{\max}(\Delta\bm{B_a}^t) \cdot \left\lVert \bm{\mu}_{\bm{a}} \right\rVert_2.
\end{align}
Based on that, we employ Lemma \ref{lem:sv_bound} to provide a high probability bound on the largest singular value, which completes the proof.
\end{IEEEproof}

\bibliographystyle{IEEEtran}
\bibliography{reference}

\end{document}